\documentclass[10pt,journal,compsoc]{IEEEtran}

\usepackage{hyperref}       
\usepackage{url}            
\usepackage{booktabs}       
\usepackage{amsfonts}       
\usepackage{nicefrac}       
\usepackage{microtype}      
\usepackage{bm}
\usepackage{wrapfig}
\usepackage{graphicx}
\usepackage{multirow}
\usepackage{colortbl}
\usepackage[dvipsnames,table]{xcolor}
\usepackage{flushend}

\usepackage{booktabs}
\usepackage{dsfont}

\usepackage{arydshln}
\usepackage{array}
\usepackage[capitalize]{cleveref}
\usepackage{siunitx}
\usepackage{amssymb}
\usepackage{pifont}
\definecolor{RowColor}{rgb}{0.95, 0.95, 1}
\definecolor{cgray}{RGB}{220,220,220}
\definecolor{cblue}{RGB}{140,190,250}
\definecolor{cpink}{RGB}{255,0,255}
\definecolor{corange}{RGB}{255,155,0}
\ifCLASSOPTIONcompsoc
  \usepackage[nocompress]{cite}
\else
  \usepackage{cite}
\fi

\newcommand{\cmark}{\textcolor{green}{\ding{51}}}%
\newcommand{\xmark}{\textcolor{red}{\ding{55}}}%

\begin{document}

\title{MsSVT++: Mixed-scale Sparse Voxel Transformer with Center Voting for 3D Object Detection}

\author{Jianan Li,
        Shaocong Dong,
        Lihe Ding,
        and Tingfa Xu 
\IEEEcompsocitemizethanks{
\IEEEcompsocthanksitem 
J. Li, S. Dong, L. Ding and T. Xu are with Beijing Institute of Technology, Beijing 10081, China, and Key Laboratory of Photoelectronic Imaging Technology and System, Ministry of Education of China, Beijing 100081,
China. \protect\\ 
E-mail: \{lijianan, ciom\_xtf1\}@bit.edu.cn
\IEEEcompsocthanksitem 
T. Xu is also with Beijing Institute of Technology Chongqing Innovation Center, Chongqing 401135, China.
\IEEEcompsocthanksitem 
A preliminary version of this work appeared at NeurIPS~\cite{dong2022mssvt}.
\IEEEcompsocthanksitem 
The source code is available at https://github.com/dscdyc/MsSVT.
}}

\markboth{Journal of \LaTeX\ Class Files,~Vol.~14, No.~8, August~2015}%
{Shell \MakeLowercase{\textit{et al.}}: Bare Demo of IEEEtran.cls for Computer Society Journals}
%

\IEEEtitleabstractindextext{%
\begin{abstract}
Accurate 3D object detection in large-scale outdoor scenes, characterized by considerable variations in object scales, necessitates features rich in both long-range and fine-grained information. While recent detectors have utilized window-based transformers to model long-range dependencies, they tend to overlook fine-grained details. To bridge this gap, we propose MsSVT++, an innovative Mixed-scale Sparse Voxel Transformer that simultaneously captures both types of information through a divide-and-conquer approach. This approach involves explicitly dividing attention heads into multiple groups, each responsible for attending to information within a specific range. The outputs of these groups are subsequently merged to obtain final mixed-scale features. To mitigate the computational complexity associated with applying a window-based transformer in 3D voxel space, we introduce a novel Chessboard Sampling strategy and implement voxel sampling and gathering operations sparsely using a hash map. Moreover, an important challenge stems from the observation that non-empty voxels are primarily located on the surface of objects, which impedes the accurate estimation of bounding boxes. To overcome this challenge, we introduce a Center Voting module that integrates newly voted voxels enriched with mixed-scale contextual information towards the centers of the objects, thereby improving precise object localization. Extensive experiments demonstrate that our single-stage detector, built upon the foundation of MsSVT++, consistently delivers exceptional performance across diverse datasets.
\end{abstract}

}

\maketitle

\IEEEraisesectionheading{\section{Introduction}\label{sec:introduction}}

\IEEEPARstart{T}{he} burgeoning interest in autonomous driving applications has brought increased attention towards 3D object detection from point cloud. However, unlike 2D images with a structured pixel arrangement, LiDAR point clouds possess inherent disorder and irregularity, posing challenges in applying CNN-like operations~\cite{he2016deep,huang2017densely} to them.
To overcome this obstacle, numerous researchers have chosen to transform point clouds into regular voxel grids~\cite{maturana2015voxnet} and leverage 3D CNNs to extract comprehensive voxel features. In recent times, inspired by the accomplishments of the vision transformer (ViT) in 2D image analysis~\cite{vaswani2017attention}, efforts have been made to extend efficient window-based transformers to 3D voxels~\cite{mao2021voxel} or pillars~\cite{fan2021embracing}.
While these approaches have demonstrated success in capturing long-range information by harnessing the powerful capabilities of transformers, they neglect the fact that indiscriminately expanding receptive fields can obscure the crucial fine-grained details necessary for accurate object recognition and localization, particularly in sparse 3D space.

\begin{figure*}
    \centering
    \centering{\includegraphics[width=0.9\textwidth]{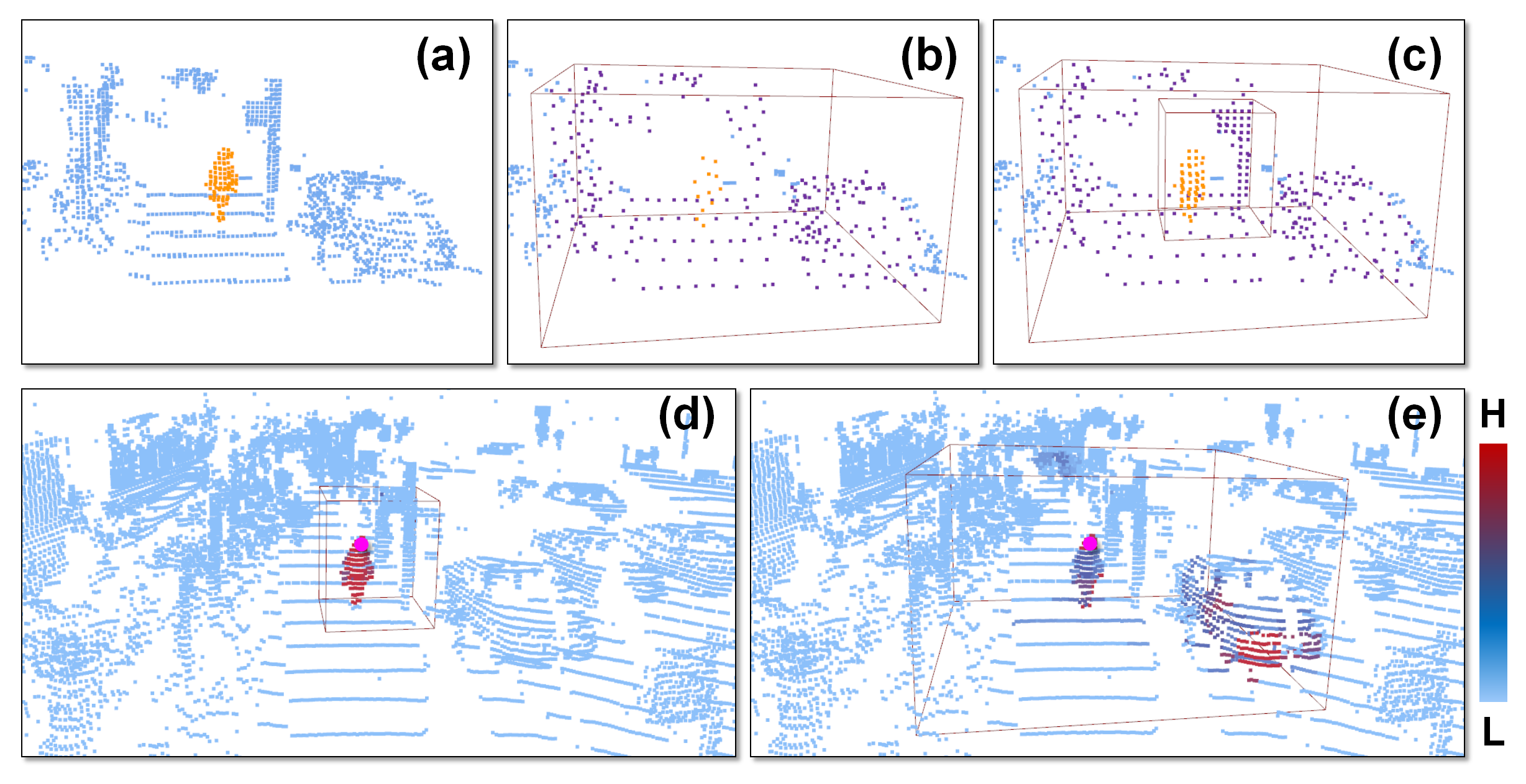}}
    \caption{\textbf{Top}: In contrast to sampling key voxels from \textbf{(b)} a single-scale 3D window in \textbf{(a)} raw point clouds, our MsSVT samples key voxels from \textbf{(c)} multi-scale windows, maintaining finer granularity on the \textcolor{corange}{target object} while covering a \textcolor{purple}{large-scale neighborhood}. \textbf{Bottom}: The different head groups in our mode accept keys sampled from windows of varying scales and are individually responsible for obtaining \textbf{(d)} fine-grained details and \textbf{(e)} long-range context, as depicted by the \textcolor{red}{higher} attention weights. This collaborative effort enables accurate object detection.}
    \label{fig:weight}
\end{figure*}

In the conventional form of window-based transformers, query features are updated within a local window by attending to keys from the same window. However, to effectively capture both long-range context and fine-grained details, it becomes necessary to enlarge the window size to encompass both local and distant voxels. Unfortunately, directly incorporating all voxels within the window as keys results in a significant increase in computational load that scales cubically with the window size.
To address this challenge, some efforts have been made to mitigate the computational burden by sampling a certain number of key voxels~\cite{mao2021voxel}. However, simplistic sampling strategies often lead to sparse sampling of local voxels (\cref{fig:weight} b), causing the model to be biased towards long-range context.
To overcome this limitation, we propose the utilization of multiple key windows of different sizes centered on a query window. Subsequently, we independently sample the same number of local and distant key voxels from the smaller and larger windows, respectively. This novel approach allows the model to preserve finer granularity in the local region and retain fine-grained details while simultaneously capturing distant voxels to expand the receptive field (\cref{fig:weight} c).

Once the voxels have been sampled, the next challenge lies in effectively attending to voxels from different windows while capturing both long-range context and fine-grained details. To tackle this issue, we propose a divide-and-conquer approach inspired by recent studies~\cite{vaswani2017attention, zhou2022understanding, park2022vision} that illustrate how transformers can learn distinct levels of self-attention using different heads.
Building upon this insight, we introduce a novel backbone network called Mixed-scale Sparse Voxel Transformer (MsSVT). The key idea is to explicitly divide the transformer heads into multiple groups, with each group dedicated to processing voxels sampled from windows of a specific size. This division allows us to capture information at different scales, effectively encompassing both long-range context and fine-grained details. By aggregating the outputs from all head groups, we achieve a comprehensive representation that incorporates mixed-scale information.
Furthermore, we devise a scale-aware relative position encoding strategy that adapts the position encoding used in each head group based on the range of the corresponding keys. This strategy ensures that the position encoding aligns with the specific scale of the voxels being processed.
We provide attention maps resulting from two different head groups in~\cref{fig:weight} (d) and (e). These maps demonstrate how the MsSVT effectively attends to voxels from different windows, capturing both local and distant dependencies.
Moreover, the mixed-scale attention in MsSVT facilitates information exchange across local windows, making it more compact compared to window-based transformers~\cite{liu2021swin,fan2021embracing} that typically require additional shift window operations.

To improve the computational efficiency of applying transformers in the 3D voxel space, we propose two strategies. Firstly, we introduce a novel approach called Chessboard Sampling (CBS) to reduce the number of query voxels that need to be sampled within the query window. CBS involves marking each voxel in the query window with one of four symbols in a chessboard-like pattern. Within each MsSVT block, we sample queries from non-empty voxels marked with a specific symbol and update their features through attention learning. For unsampled non-empty voxels, we employ a K-nearest query voxel interpolation approach, where the K-nearest updated query voxels are used to linearly interpolate the features. By employing this circular pattern of symbols, we ensure comprehensive coverage of all voxels with less computational cost.
Secondly, we exploit the sparsity of non-empty voxels by exclusively performing mixed-scale window-based attention on the non-empty sites in 3D space. Furthermore, we parallelize the search and feature gathering process for non-empty voxels using hash mapping, which further accelerates the computational process.

However, due to the fact that Lidar points are typically located on the surface of an object, the interior and center of the object are often empty. As a result, our MsSVT, which exclusively operates on non-empty voxels, heavily relies on surface voxels for bounding box prediction. This may result in incomplete coverage of the entire object, especially for larger objects. Inspired by VoteNet~\cite{qi2019deep}, we assume that filling the object centers, typically occupied by empty voxels, with some generated voxels that gathers features from different parts of the object can largely mitigate this issue.

Motivated by the aforementioned challenges, we introduce an advanced transformer network, termed MsSVT++, extending the capabilities of MsSVT. Diverging from the prior work presented by Li et al.~\cite{dong2022mssvt}, MsSVT++ incorporates a novel Center Voting module to enhance the precision of 3D object detection. The Center Voting module functions in the following manner: (i) it generates a vote point for each voxel on an object by predicting the offset from the voxel center to the corresponding object center; (ii) the generated vote point set is voxelized, resulting in clustered voxels located around the object centers; (iii) these newly generated voxels are enriched with mixed-scale contextual information through a mixed-scale attention mechanism; and (iv) the original voxel grid's object centers are filled with these new voxels, creating a merged voxel grid for subsequent processing, as previously performed. By incorporating the Center Voting module, our MsSVT++ exhibits improved accuracy in predicting bounding boxes, particularly for larger objects, surpassing the capabilities of its predecessors. This advancement contributes to the development of a more practical 3D detector.

We utilized our MsSVT++ to replace the original sparse 3D CNN backbone in SECOND~\cite{yan2018second} and developed a 3D detector. To evaluate the performance of our detector, we conducted extensive experiments on three datasets: Waymo~\cite{sun2020scalability}, KITTI~\cite{geiger2013vision}, and Argoverse 2 datasets~\cite{wilson2argoverse}. Our single-stage detector, leveraging the exceptional ability of MsSVT++ to abstract mixed-scale voxel features, exhibited remarkable performance surpassing that of state-of-the-art two-stage detectors.

Our contributions can be summarized as follows:
\begin{itemize}
\item We propose a novel Mixed-scale Sparse Voxel Transformer (MsSVT) that enables the abstraction of voxel features while considering both long-range context and fine-grained details.
\item We design an efficient Chessboard Sampling strategy that significantly reduces the computational cost of applying a voxel-based transformer in 3D space. Additionally, all operations are implemented sparsely to enhance efficiency.
\item We introduce a novel Center Voting module, expanding the capabilities of MsSVT, which gathers mixed-scale contextual information and directs it towards object centers, typically occupied by empty voxels. The resulting MsSVT++ represents a noteworthy advancement in accurate 3D detection, particularly for large objects.
\end{itemize}

\section{Related work}

\subsection{3D Object Detection on Point Clouds} 
The prevailing approach for 3D object detection on point clouds relies on either voxel-based detectors~\cite{chen2017multi, yang2018hdnet, yang2018pixor, su2015multi, li2016vehicle,ku2018joint, zhou2018voxelnet, yan2018second, wang2022rbgnet, wang2022cagroupd,du2021ago} or pillar-based detectors~\cite{lang2019pointpillars, shi2022pillarnet}. VoxelNet~\cite{zhou2018voxelnet} utilizes PointNet~\cite{qi2017pointnet, shi2022pillarnet} to aggregate features within each voxel and subsequently employ sparse 3D convolution to generate the detection outcome. SECOND~\cite{yan2018second} explores enhanced sparse convolution techniques to further improve computational speed. Pointpillar~\cite{lang2019pointpillars} transforms the point cloud into pillars, allowing the use of 2D CNNs to balance efficiency and accuracy. Rapoport-Lavie et al.~\cite{rapoport2021s} adopt the Cylindrical Coordinates to leverage the natural scanning pattern of LiDAR sensors. Chen et al.~\cite{chen2020every} further investigate the use of the Bird Eye View (BEV) and Range View (RV) in the hybrid-cylindrical-spherical voxel representation. Reconfigurable Voxels~\cite{wang2020reconfigurable} improve local neighbor searching for each voxel using a random walk scheme.

Voxel-FPN~\cite{kuang2020voxel} employs a multi-scale voxelization technique to convert the input point clouds into voxels of various sizes and applies an FPN~\cite{lin2017feature} to consolidate feature maps with different resolutions. Pillar-in-Pillar~\cite{tian2019pillar} addresses the issue of misalignment in multi-scale voxelization through a center-aligned voxelization strategy with overlapping sub-voxel partitions. In contrast, our MsSVT differs significantly from multi-scale voxelization methods by acquiring mixed-scale information from distinct windows instead of relying on multiple voxelizations, providing several advantages. Firstly, our approach preserves the fundamental unit of high-resolution voxels in each window throughout the process, while Voxel-FPN consolidates high-resolution voxel information into larger voxels, resulting in a degradation of local structural information. Additionally, our windows are significantly larger than multi-resolution voxels, allowing MsSVT to benefit from a more extensive receptive field. Secondly, our mixed-scale windows naturally overlap with each other, eliminating the misalignment problem encountered in Pillar-in-Pillar and obviating the need for additional shift window operations commonly required by window-based transformers. Thirdly, our model is more flexible and efficient as it eliminates the need for multiple voxelizations.

Two-stage detectors, as referenced in works such as~\cite{li2021lidar,shi2020pv,shi2023pv,sheng2021improving}, refine the bounding boxes generated by single-stage detectors by aggregating raw point clouds or voxel features. This refined approach has achieved state-of-the-art performance in object detection tasks.

Furthermore, the process of annotating 3D information in point clouds is known to be highly laborious, expensive, and time-consuming. Consequently, there have been notable efforts to develop approaches that can learn 3D detection with reduced annotation costs. Feng et al.~\cite{feng2023clustering} proposed a clustering-based supervised learning scheme for point cloud analysis that identifies and preserves latent data structures during representation learning. Meng et al.~\cite{meng2021towards} made an initial endeavor to train a competitive 3D object detector by utilizing weak BEV supervision in conjunction with a few precisely annotated 3D instances. ProposalContrast \cite{yin2022proposalcontrast} introduced an unsupervised point cloud pre-training framework that learns robust 3D representations by contrasting region proposals.
Moreover, Yin et al.~\cite{yin2022semi} presented a semi-supervised 3D object detection framework that enhances predictions from the teacher model through a spatial-temporal ensemble module and a clustering-based box voting module. These methods mitigate the dependency on a substantial quantity of precisely annotated samples, thereby creating new avenues for the design of 3D object detectors.

\subsection{Vision Transformer} 
The Transformer architecture~\cite{vaswani2017attention,devlin2018bert} has attained notable achievements in the field of computer vision~\cite{liu2021swin,zhu2020deformable,wang2021pyramid}. Liu et al.\cite{liu2021swin} proposed the Swin-Transformer, which imposes limitations on self-attention within non-overlapping local windows while enabling cross-window connections to enhance efficiency. Ren et al.\cite{ren2021shunted} introduced the SSA, dividing attention heads into multiple groups to aggregate image features with different granularities. Guo et al.\cite{guo2021pct} and Zhao et al.\cite{zhao2021point} have explored the application of the Transformer architecture for point cloud analysis. They have taken initial steps in adapting the Transformer for point cloud data, thereby showcasing the potential of self-attention mechanisms in capturing spatial dependencies and learning meaningful representations.

Recently, several approaches have emerged that exploit local self-attention to acquire enriched 3D feature representations~\cite{wang2022detr3d,zhang2021pvt,park2021fast,he2022voxel,mahmoud2022dense,xu2022fusionrcnn}.
Park et al.\cite{park2021fast} introduce the Fast Point Transformer, which utilizes local self-attention within a voxel hashing architecture to effectively encode continuous positional information of large-scale point clouds. PVT\cite{zhang2021pvt} combines the advantages of point-based and voxel-based representations and incorporates a sparse window attention module to alleviate computational costs.
These recent approaches leverage the Transformer architecture and self-attention to achieve enhanced performance and more comprehensive feature representations of point clouds.

In this work, we have expanded window-based attention to 3D voxels and overcome several challenges through the utilization of our innovative architecture design and technical contributions. Our work specifically addresses the following issues:
i) We successfully tackle the problem of ununiform sampling bias through our mixed-scale window design and Balanced Multi-window Sampling strategy.
ii) By incorporating scale-aware head attention, we achieve effective mixed-scale information aggregation.
iii) We significantly reduce memory and computation costs by implementing Chessboard Sampling and employing a sparse implementation approach.

\subsection{Voxel Transformer for 3D Object Detection} 
In the realm of 3D object detection, several recent methods have been proposed to enhance the performance of voxel-based transformers. VoxSeT~\cite{he2022voxel} introduced a voxel-based set attention module that conducts self-attention on voxel clusters of arbitrary size. VoTr~\cite{mao2021voxel} presented a voxel-based transformer backbone that performs self-attention on sparse voxels utilizing local and dilated attention mechanisms. 
Our work extends upon VoTr by incorporating window-based attention and optimizing sparse operations.
SWFormer~\cite{sun2022swformer} also employs sparse voxels and windows for processing 3D points, performing self-attention within each spatial window using a shift-window method akin to Swin-Transformer~\cite{liu2021swin}. In contrast, our method utilizes mixed windows and balanced sampling to achieve more efficient fusion of multi-granularity features within each attention layer. While SST~\cite{fan2021embracing}, another relevant method, follows a single-stride design and adopts the Swin-Transformer architecture, its utilization of a single window size proves inadequate for capturing multi-scale features. Consequently, SST exhibits poor performance in detecting multiple categories of varying scales simultaneously, particularly in the case of the \textit{Vehicle} category. In contrast, MsSVT overcomes this limitation by capturing mixed-scale information, thereby enhancing the detection of all categories.

\subsection{Voting Strategy for 3D Object Detection}
The integration of local neighborhood information has proven instrumental in obtaining robust and reliable estimations of object pose and dimension. As a result, several approaches have incorporated voting strategies to enhance the robustness and efficiency of 3D object detection. VoteNet~\cite{qi2019deep} introduced a pioneering end-to-end 3D object detection network that utilizes a voting mechanism inspired by the classical Hough voting approach. This mechanism allows the network to vote for object centroids from point clouds and learn to aggregate votes using their features and local geometry, thereby generating high-quality object proposals.
Building upon the foundations of VoteNet, ImVoteNet~\cite{qi2020imvotenet} integrates 2D votes from images into the 3D voting pipeline. This integration leverages existing image detectors to provide supplementary geometric and semantic information about objects. RBGNet~\cite{wang2022rbgnet} introduces a ray-based feature grouping module that encodes object surface geometry using determined rays and enhances geometric features, thereby improving detection performance.

FSD~\cite{fan2022fully} introduces an instance-wise voting module to group points into instances. While both our Center Voting module and FSD's instance-wise voting module draw inspiration from VoteNet~\cite{qi2019deep}, they differ in three key aspects: 
Firstly, FSD employs a point-based model with each input point acting as a voting seed, while MsSVT++ uses a voxel-based model with non-empty voxels as direct voting seeds, reducing computational load of the voting process due to the substantial reduction in non-empty voxels compared to all input points.
Secondly, in FSD, input points are grouped around object centers to form a grouped point cloud, upon which predictions exclusively rely. In contrast, we generate clustered voxels around object centers and integrate these voted voxels with the original voxel set from the MsSVT backbone. This preserves surface voxels while filling object centers, providing comprehensive object coverage that improves later-stage object detection.
Finally, we introduce the Mixed-scale Context Aggregation step to enhance voted voxel features through mixed-scale contextual integration. This process combines cues from different object parts, similar to the instance-wise voting module, and extends to incorporate longer-range contextual information, benefiting object detection, particularly precise object localization.

\begin{figure*}
    \centering{\includegraphics[width=0.9\textwidth]{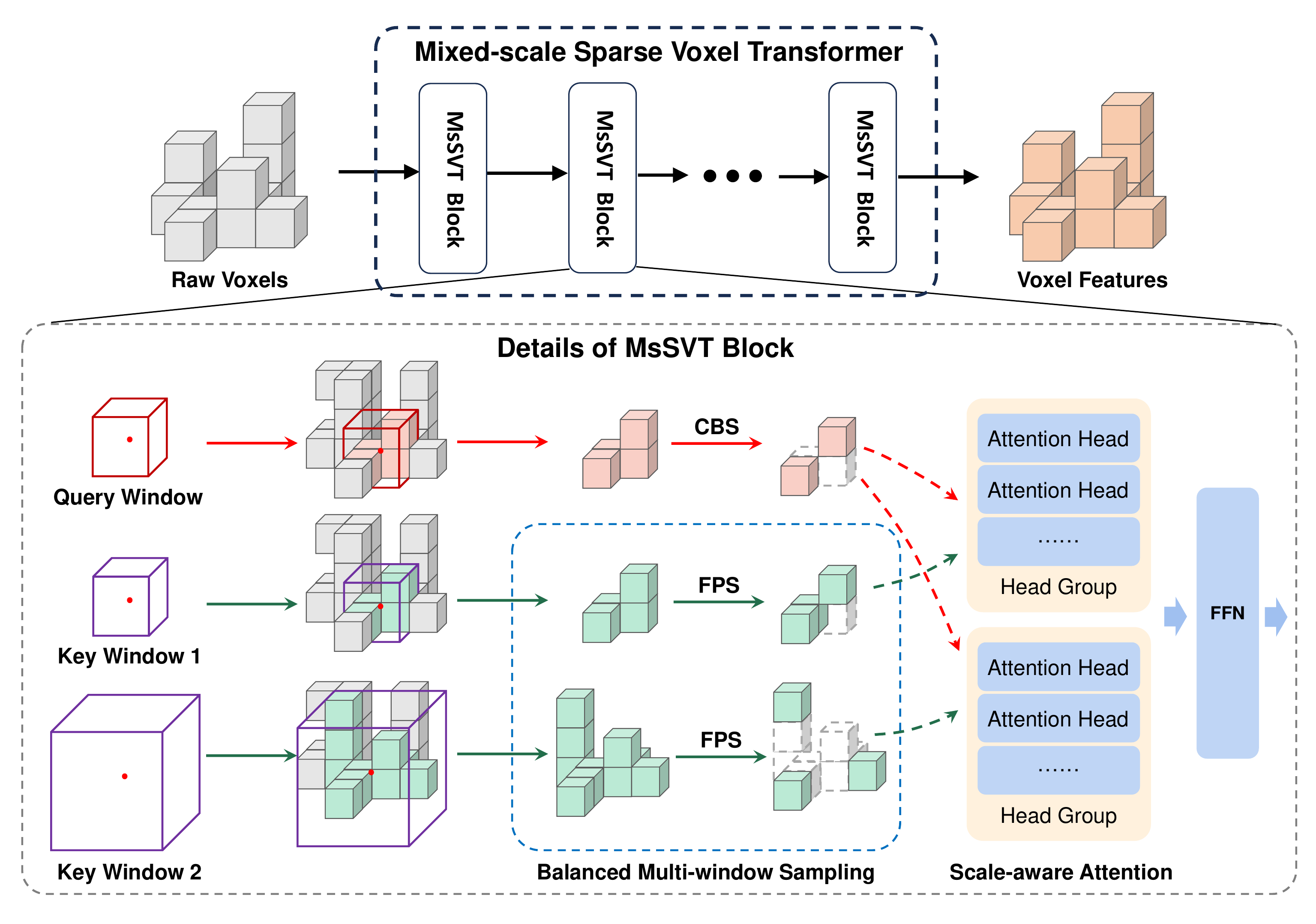} }
    \caption{Top: Architecture of our Mixed-scaled Sparse Voxel Transformer backbone. The backbone network comprises $N$ MsSVT blocks. Bottom: Implementation details of the MsSVT block. Initially, we collect non-empty voxels within the query window and employ Chessboard Sampling (CBS) to sample the queries. For the keys, we gather non-empty voxels from key windows of various sizes individually, and generate multiple sets of keys using Balanced Multi-window Sampling. Each set represents information at a specific scale. To facilitate scale-aware attention learning, keys from windows of different sizes are assigned to distinct head groups, enabling us to capture both long-range context and fine-grained details concurrently.}
     \vspace{-2mm}
    \label{fig:overall}
\end{figure*}

\section{Method} 
\label{sec:method}
In this section, we begin by providing a detailed exposition of the MsSVT block, along with its efficient sparse implementation. Subsequently, we delve into the specifics of the Center Voting module. Lastly, we present the overall construction of the 3D detector based on MsSVT++.

\subsection{Mixed-scale Sparse Voxel Transformer}
\label{sec:mix}
The overall architecture of the MsSVT block is depicted in \cref{fig:overall}. Initially, Chessboard Sampling and Balanced Multi-window Sampling techniques are employed for acquiring the query and key voxels, respectively. Subsequently, the obtained queries and keys are inputted into multiple head groups, facilitating the extraction of mixed-scale information via scale-aware attention learning. Furthermore, scale-aware relative position encoding is integrated to enhance the utilization of positional information across diverse head groups.

\subsubsection{Balanced Multi-window Sampling}
\label{sec:balanced}
Let $\bm{\mathcal{V}} = \left \{ \bm{v}_i | \bm{v}_i = (\bm{x}_i, \bm{f}_i) \right \}_{i=1}^{{\rm N}}$ denote the input set of voxels consisting of ${\rm N}$ voxels. Each voxel $i$ is characterized by its $xyz$ coordinates $\bm{x}_i \in \mathbb{Z}^{3}$ and a feature vector $\bm{f}_i \in \mathbb{R}^{\rm C}$.
Let $ \left \{\bm{r}_m | \bm{r}_m \in \mathbb{Z}^{3} \right \}_{m=0}^{\rm M} $ represent a sequence of window sizes. Here, $\bm{r}_0$ corresponds to the size of the query window, while $\bm{r}_{1,...,{\rm M}}$ correspond to the sizes of $\rm M$ consecutively larger key windows.
Initially, we partition the voxel set into non-overlapping 3D windows, each of size $\bm{r}_0$. We consider the non-empty windows as query windows, centered at $ \{\bm{c}_j | \bm{c}_j \in \mathbb{Z}^{3}\}_{j=0}^{\rm L}$, where $\rm L$ denotes the total number of query windows. The query voxels $\bm{\mathcal{V}}_{\bm{c}_j, \bm{r}_0}$ for the query window centered at $\bm{c}_j$ can be obtained by collecting ${\rm N}_q$ non-empty voxels within the window.
To ensure computational efficiency, we introduce a novel Chessboard Sampling strategy, which is comprehensively explained in~\cref{sec:sampling}.

To identify key voxels, rather than sampling within a single large window at once, as done in prior methods~\cite{mao2021voxel}, which could potentially introduce biases towards local or distant voxels, we suggest searching for neighbors of each center $\bm{c}_j$ within multiple key windows of different sizes $ \left \{\bm{r}_m | \bm{r}_m \in \mathbb{Z}^{3} \right \}_{m=1}^{\rm M} $. 
For a key window of size $\bm{r}_m$, we gather a maximum of ${\rm N}_p$ non-empty voxels within the window, represented as $\bm{\mathcal{V}}_{\bm{c}_j, \bm{r}_m}$, where ${\rm N}_p$ is a predetermined number. 
To reduce computational cost and ensure balanced sampling, we employ the farthest point sampling (FPS) algorithm to uniformly sample ${\rm N}_k$ voxels from $\bm{\mathcal{V}}_{\bm{c}_j, \bm{r}_m}$, yielding the final key voxels $\bm{\mathcal{V}}_{\bm{c}_j, \bm{r}_m}^{fps}, m=1,...,{\rm M}$ at different scales. Here, ${\rm N}_k$ denotes a pre-established maximum number of sampled voxels. By employing this multi-window strategy coupled with uniform sampling through FPS, we achieve balanced sampling of key voxels at various scales, which is essential for capturing mixed-scale information.

\subsubsection{Chessboard Sampling}
\label{sec:sampling}
To sample the keys, it is unnecessary to preserve all the key voxels. Instead, we can opt for selecting representative voxels in order to minimize computational redundancy. However, this particular approach cannot be extended to queries. Following an attention layer, it becomes crucial to retain and update each query voxel to avoid any potential loss of vital information. Nevertheless, reducing the number of queries is imperative due to the substantial rise in computational cost and memory requirements associated with larger window sizes, making their implementation impractical.

\begin{figure}[t]
    \centering{\includegraphics[width=0.46\textwidth]{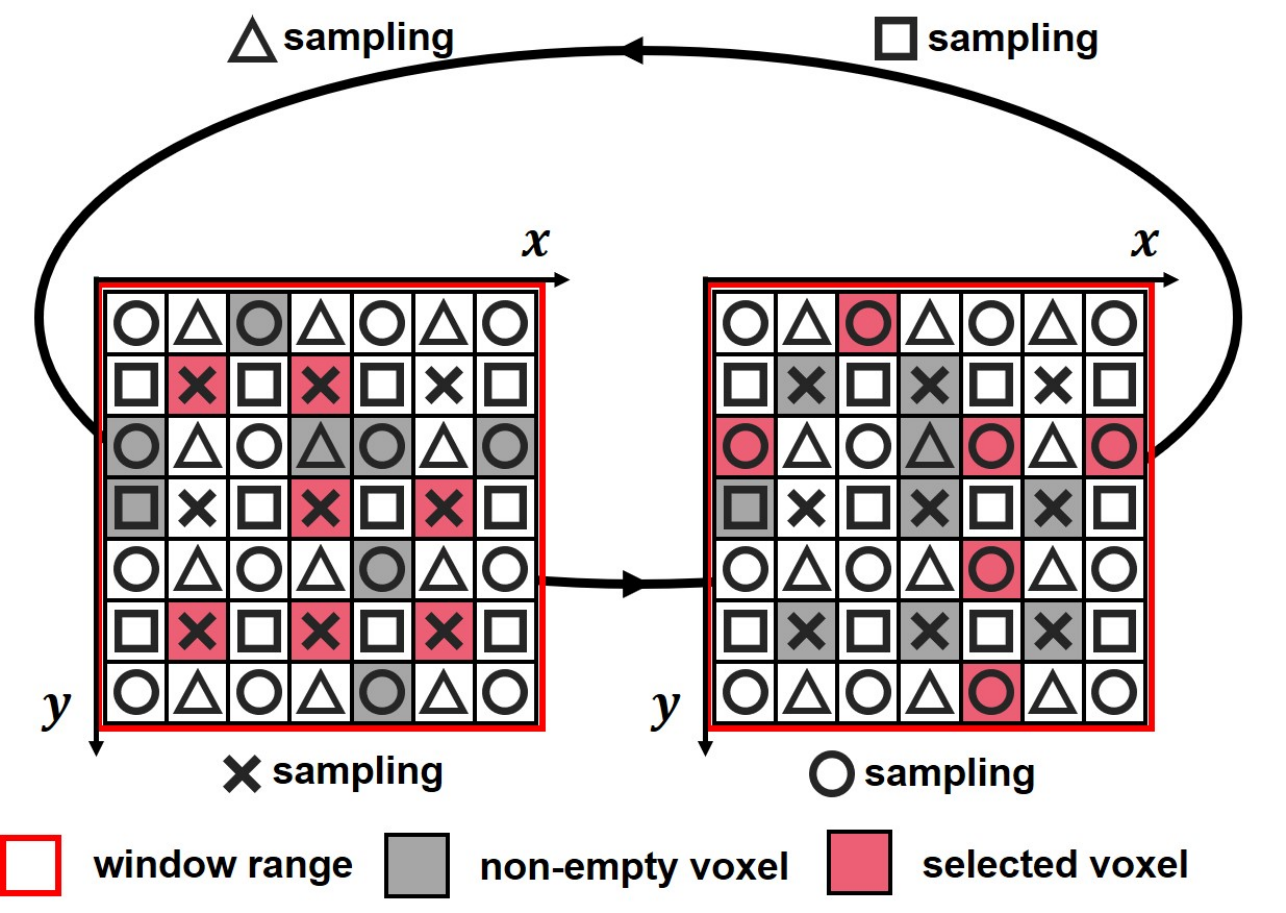}}
    \caption{Diagram of the Chessboard Sampling strategy.}
    \label{fig:cbs}
\end{figure}

Given that the positions of non-empty voxels remain unaltered throughout multiple attention blocks, we propose a solution that entails sampling a subset of query voxels for feature updating purposes. This subset is then utilized to update the unsampled query voxels. We depict this solution in \cref{fig:cbs} and refer to it as the Chessboard Sampling (CBS) strategy.

In the Chessboard Sampling strategy, each voxel within the query window is assigned one of four symbols: "$ \times $", "$ \bigcirc $", "$ \triangle $", or "$ \square $". These symbols are distributed in an evenly spaced pattern. Within each MsSVT block, we sample queries exclusively from the non-empty voxels marked with a particular symbol and update their features through attention learning. Subsequently, the features of unsampled non-empty voxels are updated by identifying the K-nearest query voxels (with a default value of ${\rm K}=3$) and linearly interpolating their updated features. The four symbols are employed in a circular pattern to sample query voxels across stacked blocks. This method enables us to preserve the original structure and encompass all voxels as comprehensively as possible. Additionally, we can apply interval sampling on any of the $x$,$y$,$z$ axes to achieve sampling rates of 1/2, 1/4, or 1/8. Typically, this technique is implemented in the horizontal $x$-$y$ plane, as depicted in \cref{fig:cbs}.

\subsubsection{Scale-aware Head Attention}
\label{sec:sh_attn}
Given the query voxels $\bm{\mathcal{V}}_{\bm{c}_j, \bm{r}_0} = (\bm{X}_0, \bm{F}_0)$ with voxel coordinates $\bm{X}_0 \in \mathbb{Z}^{{\rm N}_q \times 3}$ and feature vectors $\bm{F}_0  \in \mathbb{R}^{{\rm N}_q \times {\rm C}}$, as well as the multi-scale key voxels $\bm{\mathcal{V}}_{\bm{c}_j, \bm{r}_m}^{fps} = (\bm{X}_m, \bm{F}_m), m=1,...,{\rm M}$ with voxel coordinates $\bm{X}_m \in \mathbb{Z}^{{\rm N}_k \times 3}$ and feature vectors $\bm{F}_m  \in \mathbb{R}^{{\rm N}_k \times {\rm C}}$, we compute the queries $\bm{Q} \in \mathbb{R}^{{\rm N}_q \times {\rm C}}$, keys $\{\bm{K}_m \in \mathbb{R}^{{\rm N}_k \times {\rm C}/{\rm M}}  \}_{m=1}^{\rm M}$, and values $\{\bm{V}_m \in \mathbb{R}^{{\rm N}_k \times {\rm C}/{\rm M}} \}_{m=1}^{\rm M}$ as follows:
\begin{equation}
    \bm{Q},\ \bm{K}_m,\ \bm{V}_m = \bm{F}_0 \bm{W}^Q,\ \bm{F}_m \bm{W}_m^K,\ \bm{F}_m \bm{W}_m^V,\ \ \ m=1,...,{\rm M} 
\end{equation}
where $ \bm{W}^{Q}\in \mathbb{R}^{{\rm C} \times {\rm C}}$ and $ \bm{W}_m^{K} $,$ \bm{W}_m^{V}\in \mathbb{R}^{{\rm C} \times {\rm C}/{\rm M}}$ represent linear projections.

To facilitate scale-aware attention learning, we group the multiple attention heads into ${\rm M}$ distinct groups and assign keys from windows of various sizes to different head groups. Likewise, we divide the feature channels of the queries $\bm{Q}$ into ${\rm M}$ groups. Specifically, the $m$-th channel group of $\bm{Q}$ is denoted as $\bm{Q}_m = \bm{Q}[:, :, ~ (m-1) \times {\rm C}/{\rm M}:m \times {\rm C}/{\rm M}], m=1,...,{\rm M}$ and is fed into the corresponding $m$-th head group. This approach enables each head group to focus on learning attention patterns at a specific scale. The attended features for the $m$-th head group are represented as:
\begin{equation}
    \label{eq:attn}
    \bm{\tilde{Y}_m} = \textbf{MHA}(\bm{Q}_m, \bm{K}_m, \bm{V}_m, \textbf{RPE}(\bm{X}_0, \bm{X}_m)).
\end{equation}
Here, \textbf{MHA}$(\cdot)$ refers to the multi-head attention operation, and \textbf{RPE}$(\cdot)$ denotes the newly proposed relative position encoding, which will be elaborated in detail in~\cref{sec:relative}. Each window size corresponds to a head group with one or more attention heads.

The outputs from all head groups, denoted as $ \{\bm{\tilde{Y}}_m \in \mathbb{R}^{{\rm N}_q \times {\rm C}/{\rm M}} \}_{m=1}^{{\rm M}} $, are concatenated to form $ \bm{\tilde{Y}} \in \mathbb{R}^{{\rm N}_q \times {\rm C}}$. Subsequently, a feed-forward network (FFN) implemented with a multi-layer perceptron (MLP) is employed to process $\bm{\tilde{Y}}$ and obtain the final mixed-scale feature $\bm{Y}$. This process can be described as follows:
\begin{equation}
    \bm{\tilde{Y}} = \textbf{CAT}(\bm{\tilde{Y}}_{1}, ..., \bm{\tilde{Y}}_{{\rm M}}),
\end{equation}
\begin{equation}
    \bm{Y} = \textbf{MLP}(\textbf{LN}(\bm{\tilde{Y}})) + \bm{\tilde{Y}}. 
\end{equation}
Here, $\textbf{LN}(\cdot)$ refers to layer normalization.

\subsubsection{Scale-aware Relative Position Encoding}
\label{sec:relative}
Relative position encoding plays a vital role in transformer-based networks, as deepening the network may lead to the degradation of fine-grained position information within high-level features. To address this challenge and enhance multi-scale feature learning, we have incorporated a scale-aware adaptive relative position encoding strategy, drawing inspiration from prior studies~\cite{shaw2018self,wu2021rethinking,yang2022unified}. This strategy enables the dynamic generation of positional bias based on the distinct head groups and scales involved.

Specifically, we introduce a learnable embedding table $ \bm{T}_m \in \mathbb{R}^{{\rm C}/{\rm M} \times {\rm R}} $ for the $m$-th head group, determined by the size of the largest key window~\cite{liu2021swin}. Here, ${\rm R}$ represents the number of possible relative position pairs. The relative positional bias for the queries is computed as:
\begin{equation}
    \bm{B}_m^{Q} = \bm{\mathcal{G}}(\bm{Q}_m \bm{T}_m, \bm{I}_m)\in \mathbb{R}^{{\rm N}_q \times {\rm N}_k}. 
\end{equation}
Here, $ \bm{I}_m \in \mathbb{Z}^{{\rm N}_q \times {\rm N}_k} $ denotes the table indices that correspond to the actual relative positions between the queries and keys, while $ \bm{\mathcal{G}}(\cdot) $ represents the operation of gathering features based on the indices. Similarly, we obtain the relative positional bias for the keys, denoted as $ \bm{B}_m^{K}\in \mathbb{R}^{{\rm N}_q \times {\rm N}_k}$. Subsequently, these biases are directly incorporated into the attention weights, modifying the original attention equation in~\cref{eq:attn} as follows:
\begin{equation}
    \bm{\tilde{Y}_m} = \bm{\sigma}(\frac{\bm{Q}_m \bm{K}_m^\top}{\sqrt{{\rm C}/{\rm M}}} + \bm{B}_m^{Q} + \bm{B}_m^{K}) \bm{V}_m,
\end{equation}
where $ \bm{\sigma}(\cdot) $ represents the softmax function. Consequently, this approach allows for the adaptive adjustment of position embeddings to accommodate various scales, thereby improving the effectiveness of scale-aware head attention.

\begin{figure*}[t]
  \centering{\includegraphics[width=\textwidth]{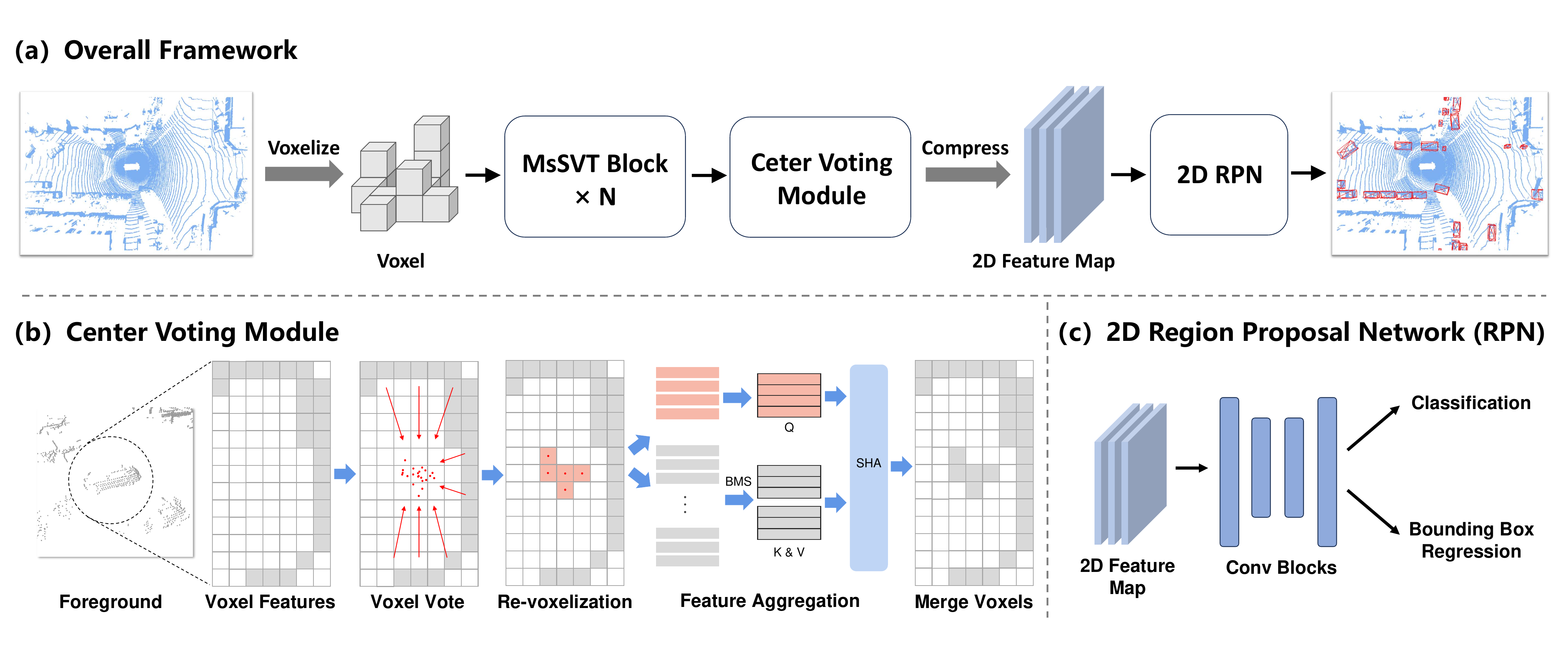}}
  \caption{\textbf{(a)} The overarching architecture of our detection framework. \textbf{(b)} Implementation details of the Object Center Voting module. This module comprises several sequential steps. Initially, it segments foreground voxels and predicts the Euclidean space offset for each foreground voxel center in relation to the object center. This predictive process generates a densely distributed vote point set in proximity to the object's center. Subsequently, the generated point set is re-voxelized, resulting in a new set of voxels. These new voxels are then enriched with mixed-scale context, incorporating information from diverse parts of the object via a mixed-scale attention mechanism. Finally, the new voxels are merged with the original voxels for further processing. Here, BMS stands for Balanced Multi-window Sampling, and SHA represents Scale-aware Head Attention. \textbf{(c)} Diagram depicting the 2D Region Proposal Network (RPN).}
  \label{fig:aggr}
  \vspace{-2mm}
\end{figure*}

\subsection{Sparse Implementation}
Running the voxel transformer directly within a 3D voxel space would result in a significant memory and computational overhead, making it impractical for successful implementation.
In order to leverage the inherent sparsity of point clouds and improve computational efficiency, we employ a sparse implementation approach for various operations involving window center searching, window gathering, and voxel sampling and gathering. These operations are implemented using CUDA operations and primarily rely on a hash map, as discussed in \cite{mao2021voxel} (\cref{sec:hash}), which establishes the mapping from the coordinate space to voxel index. For instance, during the window gathering operation, we search for each feasible position relative to the specified center within the window (\cref{sec:win_part}), and retrieve the corresponding features only if the position is a valid key in the pre-built hash map (\cref{sec:voxel_gather}). Now, we will provide a detailed explanation of how to implement window-based attention on sparse voxels.

\subsubsection{Hash Table Establishment}
\label{sec:hash}
To begin, we construct a hash table to facilitate efficient voxel searching based on the input sparse voxel set $\bm{\mathcal{V}} = \left \{ \bm{v}_i | \bm{v}_i = (\bm{x}_i, \bm{f}_i) \right \}_{i=1}^{{\rm N}}$. In this hash table, the keys are represented by the flattened voxel coordinates $ b x_{\text{max}} y_{\text{max}} z_{\text{max}} + x y_{\text{max}} z_{\text{max}} + y z_{\text{max}} + z $, while the values store the voxel features. Here, $b, x, y, z$ denote the batch index and voxel coordinates, and $ x_{\text{max}}, y_{\text{max}}, z_{\text{max}} $ represent the maximum range of the voxel space. The hash function employs a modulus strategy and employs linear probing to locate an empty address in the event of a hash collision. This approach allows us to only retain non-empty voxels and establish a connection between voxel coordinates and their corresponding features. By providing a voxel coordinate, we can efficiently determine if it is empty and retrieve the associated feature data for non-empty locations. It is worth noting that the CUDA parallel processing capability enables the hash table to be processed in parallel, thereby further accelerating the voxel searching process.

\subsubsection{Sparse Window Partition}
\label{sec:win_part}
In \cref{sec:balanced}, the input voxel set is partitioned into non-overlapping 3D windows based on a specified window size $ \bm{r}_{0} $. Subsequently, we identify the non-empty windows as query windows and determine their corresponding window centers $ \{\bm{c}_j | \bm{c}_j \in \mathbb{Z}^{3}\}_{j=0}^{{\rm L}} $ using the following formula:
\begin{equation}
    \bm{c}_j = ( \lfloor \bm{x}_j / \bm{r}_{0} \rfloor + 0.5) \times \bm{r}_{0},
\end{equation}
Here, $\bm{x}_j$ represents the $xyz$ coordinates of the central voxel of the $j$-th query window.
Notably, we take further steps to eliminate any duplicate centers, resulting in a unique set of centers. This process contributes to reducing computational costs.

\subsubsection{Voxel Sampling and Gathering}
\label{sec:voxel_gather}
Once we have determined the window centers, our next step is to search for non-empty voxels surrounding these centers within either the query or key windows. By leveraging the pre-built hash table, the voxel search process can be transformed into locating existing hash keys using the hash function. This enables us to identify the coordinates of non-empty voxels within all non-empty windows. Subsequently, we employ Balanced Multi-window Sampling to obtain the final set of sampled voxel coordinates. Finally, we efficiently gather the voxel features from the hash table by utilizing the sampled voxel coordinates. These voxel coordinates, along with their corresponding features, are then fed into the attention mechanism described in \cref{sec:sh_attn}.

\subsection{Center Voting Module}
\label{sec:aggregation}
Let $\bm{\mathcal{V}}' = \left \{ \bm{v}_i' | \bm{v}_i' = (\bm{x}_i, \bm{f}_i') \right \}_{i=1}^{{\rm N}}$ denote the refined voxel set obtained from the MsSVT backbone, where $\bm{x}_i \in \mathbb{R}^3$ represents the center of the voxel and $\bm{f}_i' \in \mathbb{R}^{\rm C}$ represents the refined voxel feature for the $i$-th voxel. The Center Voting module aims to produce a new voxel set $\bm{\mathcal{V}}'' = \left \{ \bm{v}_i'' | \bm{v}_i'' = (\bm{x}_i, \bm{f}_i'') \right \}_{i=1}^{{\rm N}}$, where the initially empty voxels near the center of the object are replaced with non-empty voxels containing aggregated features from various parts of the objects. The overall process, as illustrated in~\cref{fig:aggr} (b), consists of four main stages: Vote generation, Vote set re-voxelization, Mixed-scale context aggregation, and Voxel merging.

\noindent\textbf{Vote Generation.} 
Each non-empty voxel in $\bm{\mathcal{V}}'$ is treated as a seed voxel, and a shared voting module independently generates a vote point from each seed voxel. The seed feature $\bm{f}_i'$ is used as input to the voting module, which employs a multi-layer perceptron (MLP) network to learn a mapping function $\bm{\phi}(\cdot)$, producing the Euclidean space offset $\Delta \bm{x}_i \in \mathbb{R}^3$ according to:
\begin{equation}
    \Delta \bm{x}_i = \bm{\phi} \left ( \bm{f}_i' \right ),
\end{equation}
The resulting vote point generated from the seed voxel $\bm{v}_i'$ is represented as $(\bm{x}_i+\Delta \bm{x}_i, \bm{f}_i')$.
To supervise the predicted offset $\Delta \bm{x}_i$, a regression loss is employed, defined as follows:
\begin{equation}
    \mathcal{L}_{vote} = \frac{1}{{\rm N}_f} \sum_{i}\left \| \Delta \bm{x}_i - \Delta \bm{x}_i^* \right \| \mathds{1}[\bm{v}_i' \text{ is FG}],
\end{equation}
where $\mathds{1}[\bm{v}_i' \text{ is FG}]$ indicates whether a seed voxel is a foreground voxel. Foreground voxels are those that fall within a ground-truth bounding box, typically located on an object's surface. ${\rm N}_f$ denotes the total number of foreground voxels. $\Delta \bm{x}_i^*$ represents the ground-truth displacement from the seed position $\bm{x}_i$ to the bounding box center of the corresponding object.

Simultaneously, we train the model to predict objectness scores for each voxel, given by:
\begin{equation}
     p_i = \bm{\Phi} \left ( \bm{f}_i' \right ),
\end{equation}
where $\bm{\Phi}(\cdot)$ is a mapping function learned by an MLP network followed by a Sigmoid activation. The predicted objectness score $p_i$ is supervised using the Focal Loss~\cite{lin2017focal}, with the ground-truth objectness score of foreground voxels set to 1 and the rest of the voxels set to 0. During inference, the predicted objectness score is utilized to distinguish foreground voxels from background voxels.

\noindent\textbf{Vote Set Re-voxelization.}
After collecting the vote points generated from all foreground seeds, we obtain a vote point set $\left \{ (\bm{x}_i+\Delta \bm{x}_i, \bm{f}_i') \right \}_{i=1}^{{\rm N}_f}$ consisting of ${\rm N}_f$ vote points. To voxelize this set, we use the same voxel size as $\bm{\mathcal{V}}'$ and initialize the voxel feature of each non-empty voxel by computing the average of the features of all vote points falling within that voxel. Consequently, we obtain a new voxel set $\bm{\mathcal{S}} = \left \{ \bm{s}_i | \bm{s}_i = (\bm{x}_i, \bm{g}_i) \right \}_{i=1}^{{\rm N}}$. The non-empty voxels in $\bm{\mathcal{S}}$ are typically concentrated in the centers of objects.

\noindent\textbf{Mixed-scale Context Aggregation.}
The aim of this stage is to enhance the voxel feature in $\bm{\mathcal{S}}$ by integrating contextual information from different scales, incorporating cues from various object parts as well as longer-range context, through a mixed-scale attention mechanism. To achieve this, we initially divide $\bm{\mathcal{S}}$ into non-overlapping query windows. For each query window, we gather multiple key windows of different sizes from $\bm{\mathcal{V}}'$. The non-empty voxels within the query window serve as queries without Chessboard Sampling, while the corresponding multiple key windows provide the keys using a Balanced Multi-window Sampling strategy. Subsequently, the mixed-scale attention is performed, as implemented in the MsSVT block. This results in an updated voxel set $\bm{\mathcal{S}}' = \left \{ \bm{s}_i' | \bm{s}_i' = (\bm{x}_i, \bm{g}_i') \right \}_{i=1}^{{\rm N}}$, where the feature of each non-empty voxel is enriched with context from multiple scales.

It is important to note that the non-empty voxels (queries) in $\bm{\mathcal{S}}$ are primarily concentrated near the object centers and typically represent a small number. This characteristic ensures that the context aggregation step remains highly computationally efficient.

\noindent\textbf{Voxel Merging.}
In this step, the voxel set $\bm{\mathcal{S}}'$ is merged with $\bm{\mathcal{V}}'$ by mapping the features of non-empty voxels in $\bm{\mathcal{S}}'$ to the corresponding voxels in $\bm{\mathcal{V}}'$ at the same locations. Typically, the non-empty voxels in $\bm{\mathcal{S}}'$ are concentrated around object centers, whereas the voxels in $\bm{\mathcal{V}}'$ are usually empty at these locations. By merging these voxel sets, the features of empty voxels in the object centers of $\bm{\mathcal{V}}'$ are filled with informative multi-scale context features from the surrounding objects. The resulting voxel set is denoted as $\bm{\mathcal{V}}'' = \left \{ \bm{v}_i'' | \bm{v}_i'' = (\bm{x}_i, \bm{f}_i'') \right \}_{i=1}^{{\rm N}}$ and is utilized in subsequent processes as before.

\subsection{Detector Establishment}
\label{sec:detector}
Our MsSVT backbone comprises multiple MsSVT blocks, forming the foundation of our architecture. Following the traditional 3D detection approach introduced in OpenPCDet~\cite{openpcdet2020}, our single-stage detector, built upon MsSVT++, consists of a Voxel Feature Encoding (VFE) layer, a MsSVT backbone, a Center Voting module, and a 2D Region Proposal Network (RPN), as depicted by~\cref{fig:aggr}. Specifically, we replace the 3D backbone of SECOND~\cite{yan2018second} with MsSVT and integrate a Center Voting module, while maintaining the remaining network components unchanged. As MsSVT is proficient in capturing features at varying scales, there is no need for downsampling processes. The input point cloud is converted into regular voxels and fed into our MsSVT backbone, followed by the Center Voting module, to extract mixed-scale voxel features. These features are then compressed vertically using an additional MsSVT block, where the query and key window sizes in this block are set as $(1,1,\infty)$. This setting effectively compresses the 3D voxels into a 2D feature map. Specifically, the query represents the average voxel features within the pillar window. The compressed features are subsequently passed to the 2D RPN and detection head to obtain detection results. To further enhance the detection performance of our single-stage detector, we have incorporated a Region of Interest (ROI) head implemented by CT3D~\cite{sheng2021improving}. This integration has led to the development of a two-stage detector based on MsSVT++.

\begin{table*}
\caption{Results on the WOD validation set (train with $\bm{20\%}$ of the Waymo data). SS: Single-stage model, TS: Two-stage model, SF: Single frame input. It is important to note that some priors only report results of single-class training, which is generally simpler than multi-class training.}
\label{tab:waymo}
\renewcommand{\arraystretch}{1.15}
\setlength{\tabcolsep}{4.4pt}
\centering
\begin{tabular}{l|c|cc|cc|cc|cc|cc|cc}
    \hline
    \hline
    \multirow{2}{*}{Method} &\multirow{2}{*}{Reference} & \multicolumn{2}{c|}{Vel\_L1} &  \multicolumn{2}{c|}{Vel\_L2} & \multicolumn{2}{c|}{Ped\_L1} & \multicolumn{2}{c|}{Ped\_L2} & \multicolumn{2}{c|}{Cyc\_L1} & \multicolumn{2}{c}{Cyc\_L2} \\ 
    & & \multicolumn{1}{c}{mAP} & mAPH & \multicolumn{1}{c}{mAP} & mAPH & \multicolumn{1}{c}{mAP} & mAPH & \multicolumn{1}{c}{mAP} & mAPH & \multicolumn{1}{c}{mAP} & mAPH & \multicolumn{1}{c}{mAP} & mAPH \\
    \hline
    \multicolumn{14}{c}{\bf{Single-Stage Methods}}\\
    \hline
    SECOND~\cite{yan2018second} & Sensors 2018 & 70.96 & 70.34 & 62.58 & 62.02 & 65.23 & 54.24 & 57.22 & 47.49 & 57.13 & 55.62 & 54.97 & 53.53 \\
    PointPillar~\cite{lang2019pointpillars} & CVPR 2019 & 70.43 & 69.83 & 62.18 & 61.64 & 66.21 & 46.32 & 58.18 & 40.64 & 55.26 & 51.75 & 53.18 & 49.80 \\
    CenterPoint~\cite{yin2021center} & CVPR 2021 & 72.76 & 72.23 & 64.91 & 64.42 & 74.19 & 67.96 & 66.03 & 60.34  & 71.04 & 69.79 & 68.49 & 67.28 \\
    VOTR-SS~\cite{mao2021voxel} & ICCV 2021 & 68.99 & 68.39 & 60.22 & 59.69 & -- & -- & -- & -- & -- & -- & -- & -- \\
    RSN-SF~\cite{sun2021rsn} & CVPR 2021 & 75.10 & 74.60 & 66.00 & 65.50 & -- & -- & -- & -- & -- & -- & -- & -- \\
    MsSVT (SS)~\cite{dong2022mssvt} & NeurIPS 2022 & 77.18 & 76.67 & 68.75 & 68.28 & 80.25 & 73.05 & 72.88 & 66.14 & 73.75 & 72.53 & 70.96 & 69.79 \\ 
    \rowcolor{RowColor}  MsSVT++ (SS) & --& \textbf{78.53} & \textbf{77.94} & \textbf{69.88} & \textbf{69.42} & \textbf{80.63} & \textbf{73.31} & \textbf{73.02} & \textbf{66.29} & \textbf{75.47} & \textbf{74.30} & \textbf{72.48} & \textbf{71.22} \\  
    \hline
    \multicolumn{14}{c}{\bf{Two-Stage Methods}}\\
    \hline
    Part-A2~\cite{shi2019part} &TPAMI 2020 & 74.66 & 74.12 & 65.82 & 65.32 & 71.71 & 62.24 & 62.46 & 54.06 & 66.53 & 65.18 & 64.05 & 62.75 \\
    PV-RCNN~\cite{shi2020pv} &CVPR 2020 & 75.95 & 75.43 & 68.02 & 67.54 & 75.94 & 69.40 & 67.66 & 61.62 & 70.18 & 68.98 & 67.73 & 66.57 \\
    Voxel-RCNN~\cite{deng2020voxel} & AAAI 2021 & 76.13 & 75.66 & 68.18 & 67.74 & 78.20 & 71.98 & 69.29 & 63.59 & 70.75 & 69.68 & 68.25 & 67.21 \\
    VOTR-TS~\cite{mao2021voxel} & ICCV 2021& 74.95 & 74.25 & 65.91 & 65.29 & -- & -- & -- & -- & -- & -- & -- & -- \\
    LidarRCNN(2x)~\cite{li2021lidar} & CVPR 2021 & 73.5 & 73.0 & 64.7 & 64.2 & 71.2 & 58.7 & 63.1 & 51.7 & 68.6 & 66.9 & 66.1 & 64.4 \\
    PV-RCNN++~\cite{shi2023pv} & IJCV 2023 & 77.61 & 77.14 & 69.18 & 68.75 & 79.42 & 73.31 & 70.88 & 65.21 & 72.50 & 71.39 & 69.84 & 68.77 \\
    CT3D~\cite{sheng2021improving}& ICCV 2021 & 76.30 & -- & 69.04 & -- & -- & -- & -- & -- & -- & -- & -- & -- \\
    MsSVT (CT3D)~\cite{dong2022mssvt} & NeurIPS 2022 & 78.41 & 77.91 & 69.74 & 69.17 & 82.34 & 76.77 & 74.71 & 69.36 & 75.74 & 74.65 & 73.72 & 72.64 \\ 
    \rowcolor{RowColor} MsSVT++ (LidarRCNN) & -- & 78.62  & 78.04  & 69.87  & 69.33  & 82.26  & 76.59  & 74.68  & 69.31  & 75.58  & 74.49  & 73.57  & 72.50  \\
    \rowcolor{RowColor} MsSVT++ (CT3D) & -- & \textbf{79.24} & \textbf{78.63} & \textbf{70.38} & \textbf{69.72} & \textbf{82.46} & \textbf{76.84} & \textbf{74.90} & \textbf{69.51} & \textbf{76.49} & \textbf{75.32} & \textbf{74.25} & \textbf{73.18} \\ 
    \hline
    \hline
\end{tabular}
\end{table*}

\begin{table*}[]
\caption{Results on the WOD validation set (train with $\bm{100\%}$ of the Waymo data). $\dag$: Multimodal model.}
\label{tab:waymo_all}
\renewcommand{\arraystretch}{1.15}
\setlength{\tabcolsep}{4.7pt}
\centering
\begin{tabular}{l|c|cc|cc|cc|cc|cc|cc}
    \hline
    \hline
    \multirow{2}{*}{Method} &\multirow{2}{*}{Reference} & \multicolumn{2}{c|}{Vel\_L1} &  \multicolumn{2}{c|}{Vel\_L2} & \multicolumn{2}{c|}{Ped\_L1} & \multicolumn{2}{c|}{Ped\_L2} & \multicolumn{2}{c|}{Cyc\_L1} & \multicolumn{2}{c}{Cyc\_L2} \\ 
    & & \multicolumn{1}{c}{mAP} & mAPH & \multicolumn{1}{c}{mAP} & mAPH & \multicolumn{1}{c}{mAP} & mAPH & \multicolumn{1}{c}{mAP} & mAPH & \multicolumn{1}{c}{mAP} & mAPH & \multicolumn{1}{c}{mAP} & mAPH \\
    \hline
    \multicolumn{14}{c}{\bf{Single-Stage Methods}}\\
    \hline
    SECOND~\cite{yan2018second}& Sensors 2018 & 72.27 & 71.69 & 63.85 & 63.33 & 68.70 & 58.18 & 60.72 & 51.31 & 60.62 & 59.28 & 58.34 & 57.05 \\
    PointPillar~\cite{lang2019pointpillars}& CVPR 2019 & 71.57 & 70.99 & 63.06 & 62.54 & 70.61 & 56.70 & 62.85 & 50.24 & 64.36 & 62.27 & 61.95 & 59.93 \\
    PDV~\cite{hu2022point} & CVPR 2022 & 76.85 & 76.33 & 69.30 & 68.81 & 74.19 & 65.96 & 65.85 & 58.28 & 68.71 & 67.55 & 66.49 & 65.36 \\
    SST-SS-SF~\cite{fan2021embracing}& CVPR 2022 & 75.13 & 74.64 & 66.61 & 66.17 & 80.07 & 72.12 & 72.38 & 65.01 & 71.49 & 70.20 & 68.85 & 67.61 \\
    CenterFormer-SF~\cite{zhou2022centerformer} & ECCV 2022 & -- & -- & 69.70 & -- & -- & -- & 68.30 & -- & -- & -- & 68.80 & -- \\
    MsSVT (SS)~\cite{dong2022mssvt} & NeurIPS 2022 & 77.83 & 77.32 & 69.53 & 69.06 & 80.39 & 73.61 & 73.00 & 66.65 & 75.17 & 73.99 & 72.37 & 71.24 \\
    \rowcolor{RowColor} MsSVT++ (SS) &-- & \textbf{78.96} & \textbf{78.39} & \textbf{70.57} & \textbf{70.01} & \textbf{80.64} & \textbf{73.78} & \textbf{73.12} & \textbf{66.80} & \textbf{75.98} & \textbf{74.73} & \textbf{72.97} & \textbf{71.82} \\
    \hline
    \multicolumn{14}{c}{\bf{Two-Stage Methods}}\\
    \hline
    Part-A2~\cite{shi2019part}& TPAMI 2020 & 77.05 & 76.51 & 68.47 & 67.97 & 75.24 & 66.87 & 66.18 & 58.62 & 68.60 & 67.36 & 66.13 & 64.93 \\
    PV-RCNN~\cite{shi2020pv}& CVPR 2020 & 78.00 & 77.50 & 69.43 & 68.98 & 79.21 & 73.03 & 70.42 & 64.72 & 71.46 & 70.27 & 68.95 & 67.79 \\
    Graph-RCNN~\cite{yang2022graph} & ECCV 2022 & \textbf{80.77} & \textbf{80.28} & \textbf{72.55} & \textbf{72.10} & 82.35 & 76.64 & 74.44 & 69.02 & 75.28 & 74.21 & 72.52 & 71.49 \\
    PV-RCNN++~\cite{shi2023pv}& IJCV 2023 & 79.25 & 78.78 & 70.61 & 70.18 & 81.83 & 76.28 & 73.17 & 68.00 & 73.72 & 72.66 & 71.21 & 70.19 \\
     LoGoNet-SF$\dag$~\cite{li2023logonet} & CVPR 2023 & 78.95 & 78.41 & 71.21 & 70.71 & \textbf{82.92} & 77.13 & \textbf{75.49} & \textbf{69.94} & 76.61 & 75.53 & 74.53 & 73.48 \\
    MsSVT (CT3D)~\cite{dong2022mssvt} & NeurIPS 2022 & 79.35 & 78.86 & 70.65 & 70.23 & 82.41 & 77.04 & 74.74 & 69.57 & 77.12 & 76.01 & 74.98 & 74.07 \\
    \rowcolor{RowColor} MsSVT++ (CT3D) & -- & 79.96 & 79.43 & 71.30 & 70.86 & 82.49 & \textbf{77.21} & 74.78 & 69.68 & \textbf{77.46} & \textbf{76.35} & \textbf{75.29} & \textbf{74.31} \\ 
    \hline
    \hline
\end{tabular}
\end{table*}

\begin{table}  
    \caption{Runtime per scan on the WOD validation set, tested using a single Tesla V100 GPU.} 
    \label{table:runtime}
    \renewcommand{\arraystretch}{1.15}
    \setlength{\tabcolsep}{2.9pt}
    \centering
    \begin{tabular}{l|c c c c  >{\columncolor{RowColor}}c}
    \toprule
    \multirow{2}{*}{Method} & VOTR-SS & PDV & SST-SS-SF  & PV-RCNN++ & MsSVT++  \\
    & ~\cite{mao2021voxel} & ~\cite{hu2022point} & ~\cite{fan2021embracing} & ~\cite{shi2023pv}  & (SS) \\ 
    \midrule
    Runtime (ms) & 306 & 340 & 81 & 125 & 97 \\ 
    \bottomrule
    \end{tabular}
    \vspace{-1.5mm}
\end{table}

\section{Experiments} 
\label{Experiments}
In this section, we present the architectural details of MsSVT. Subsequently, we compare our model with recent state-of-the-art detectors on three different datasets, namely Waymo Open~\cite{sun2020scalability}, KITTI~\cite{geiger2013vision}, and Argoverse 2~\cite{wilson2argoverse}.

\subsection{Architectural Details}
\label{sec:imple}
MsSVT is composed of four regular MsSVT blocks, each with a query window size of (3, 3, 5) and key window sizes of (3, 3, 5) and (7, 7, 7). These blocks are followed by a Center Voting module and a specialized MsSVT block, where the windows are set as $1 \times 1$ pillar as mentioned in~\cref{sec:detector}. We divide the 8 attention heads into 2 head groups. The sampling rate of the Chessboard Sampling is 1/4, and the maximum number of sampled keys ${\rm N}_k$ is 32. We use the center head~\cite{yin2021center} to generate single-stage bounding boxes. Additionally, we provide a two-stage version incorporating CT3D~\cite{sheng2021improving}. For more details on the experimental setup, please refer to OpenPCDet~\cite{openpcdet2020}, as all our experiments are conducted using this toolbox.

\begin{figure*}[t]
    \centering{\includegraphics[width=\textwidth]{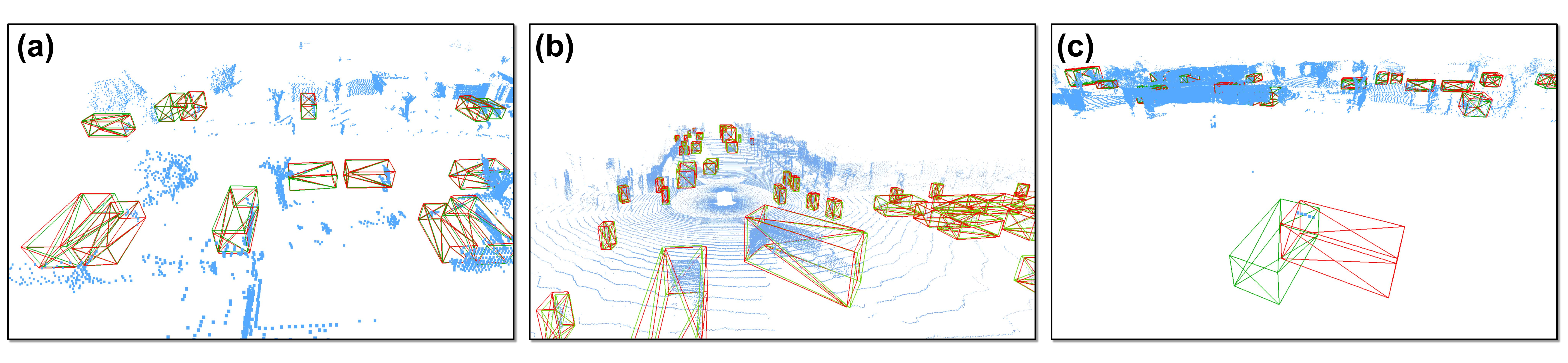}}
    \caption{Qualitative results on Waymo. Ground-truths and predictions are depicted by \textcolor{red}{red} and \textcolor{green}{green} boxes, respectively. Our method exhibits remarkable performance in scenes (a) that exceed a range of $50\rm{m}$, (b) featuring dense objects with significant scale variations, but occasionally encounters challenges in scenes (c) that encompass distant, isolated objects.}
    \label{fig:vis_box}
\end{figure*}

\begin{figure*}[t]
    \centering{\includegraphics[width=0.9\textwidth]{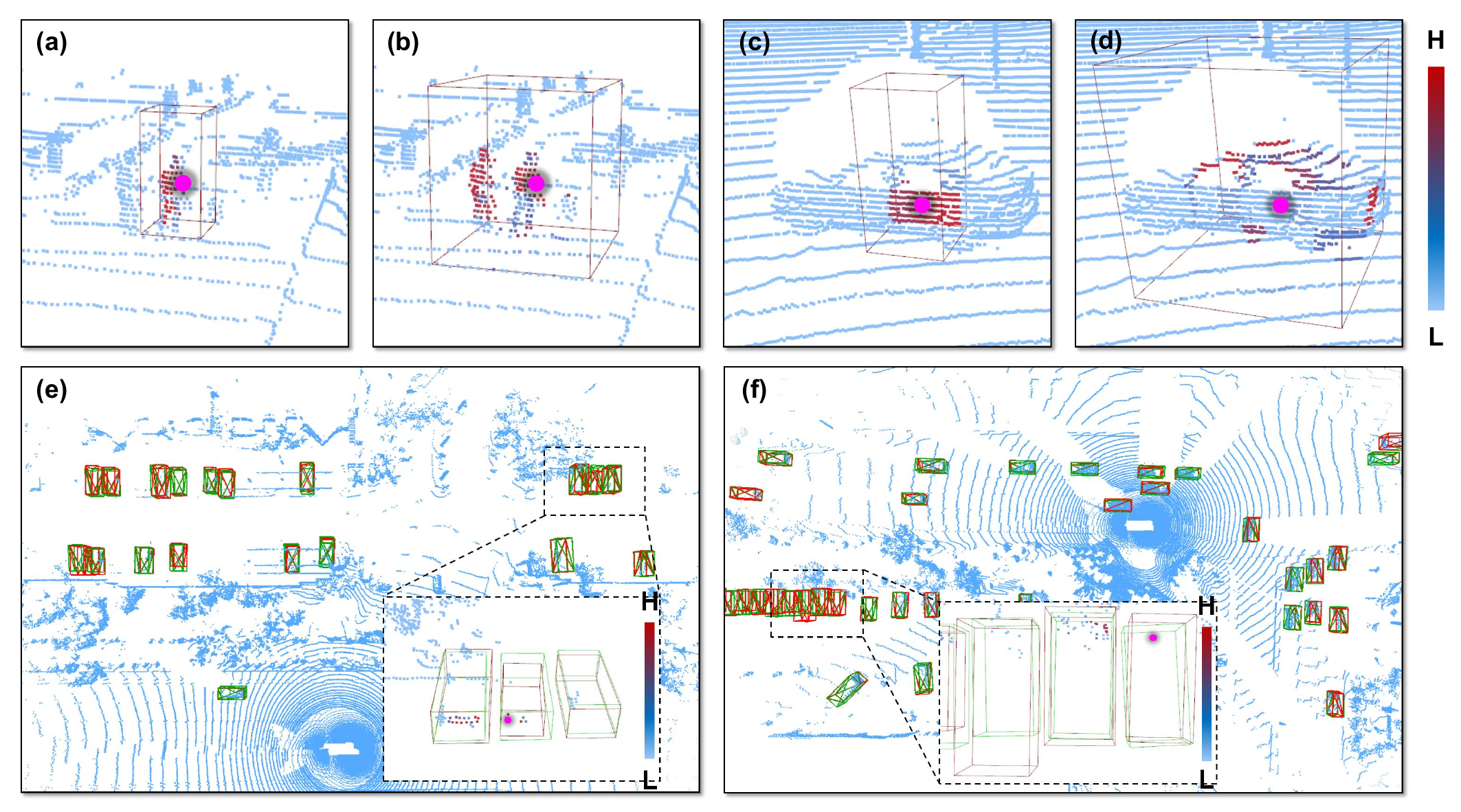}}
    \caption{\textbf{Top:} Visualization of attention maps. The \textcolor{cpink}{pink} dot denotes the query position. Positions with high and low attention weights are highlighted in \textcolor{red}{red} and \textcolor{blue}{blue}, respectively. \textbf{Bottom:} Predicted results in challenging cases on Waymo. The \textcolor{red}{red} and \textcolor{green}{green} boxes represent the ground-truth and predictions, respectively. The \textcolor{cpink}{pink} dot represents the query position. The attention weight is visually represented by the color of the point.}
    \label{fig:vis_weight}
\end{figure*}

\subsection{Results on Waymo}
\label{sec:waymo}

\noindent\textbf{Setups.}
We initially assess the performance of our model on the large-scale Waymo Open Dataset~\cite{sun2020scalability}. The input is a single-frame point cloud, covering a detection range of $150\rm{m}$$\times$$150\rm{m}$. We set the detection range as $\left[-75.2\rm{m}, 75.2\rm{m} \right]$ in the horizontal direction and $\left[-2.0\rm{m}, 4.0\rm{m} \right]$ in the vertical direction. The voxel size is set to $\left(0.4\rm{m}, 0.4\rm{m}, 0.6\rm{m}\right)$. We adopt the same training strategy as in~\cite{mao2021voxel}, whereby the model is trained using the Adam optimizer~\cite{kingma2014adam} for $80$ epochs, utilizing $20\%$ of the Waymo data. We employ the cyclic decay scheme~\cite{yan2018second}, whereby the learning rate is increased from $1\text{e-}4$ to $1\text{e-}3$ during the first $40\%$ epochs and then decreased to $1\text{e-}5$ for the remaining epochs. Additionally, we report the results obtained by training for $30$ epochs on $100\%$ of the Waymo data using the same optimizer and learning rate scheme. We assess the performance of the model using the 3D mean Average Precision (mAP) evaluation metric for difficulty levels of LEVEL 1 and LEVEL 2.

\noindent\textbf{Main Results.}
In~\cref{tab:waymo}, we compare our model with state-of-the-art priors. Notably, our model exhibits the capability to simultaneously detect three object categories, which poses a greater challenge compared to detecting a single category. Using only $20\%$ of the training data, our single-stage detector, denoted as MsSVT++ (SS), significantly outperforms other single-stage counterparts, even though some of them are specifically trained for one particular category. Remarkably, MsSVT++ (SS) achieves mAP scores of $78.53$, $80.63$, and $75.47$ for \textit{Vehicle}, \textit{Pedestrian}, and \textit{Cyclist}, respectively, even surpassing the state-of-the-art two-stage PV-RCNN++\cite{shi2023pv} by a considerable margin of $0.9$, $1.2$, and $3.0$ for the respective categories.

Similarly, Table~\ref{tab:waymo_all} demonstrates that, with the utilization of $100\%$ of the data, our method exhibits superior performance compared to the counterparts. MsSVT++ (SS) exhibits significant performance gains over the transformer-based detector SST~\cite{fan2021embracing} dedicated to small object detection, achieving a large margin of $\bm{3.8}$ and $\bm{4.5}$ mAP on \textit{Vehicle} and \textit{Cyclist}, respectively. These results clearly demonstrate the superiority of our method in capturing mixed-scale information over conventional transformer designs.

\begin{figure*}[t]
  \centering{\includegraphics[width=0.98\textwidth]{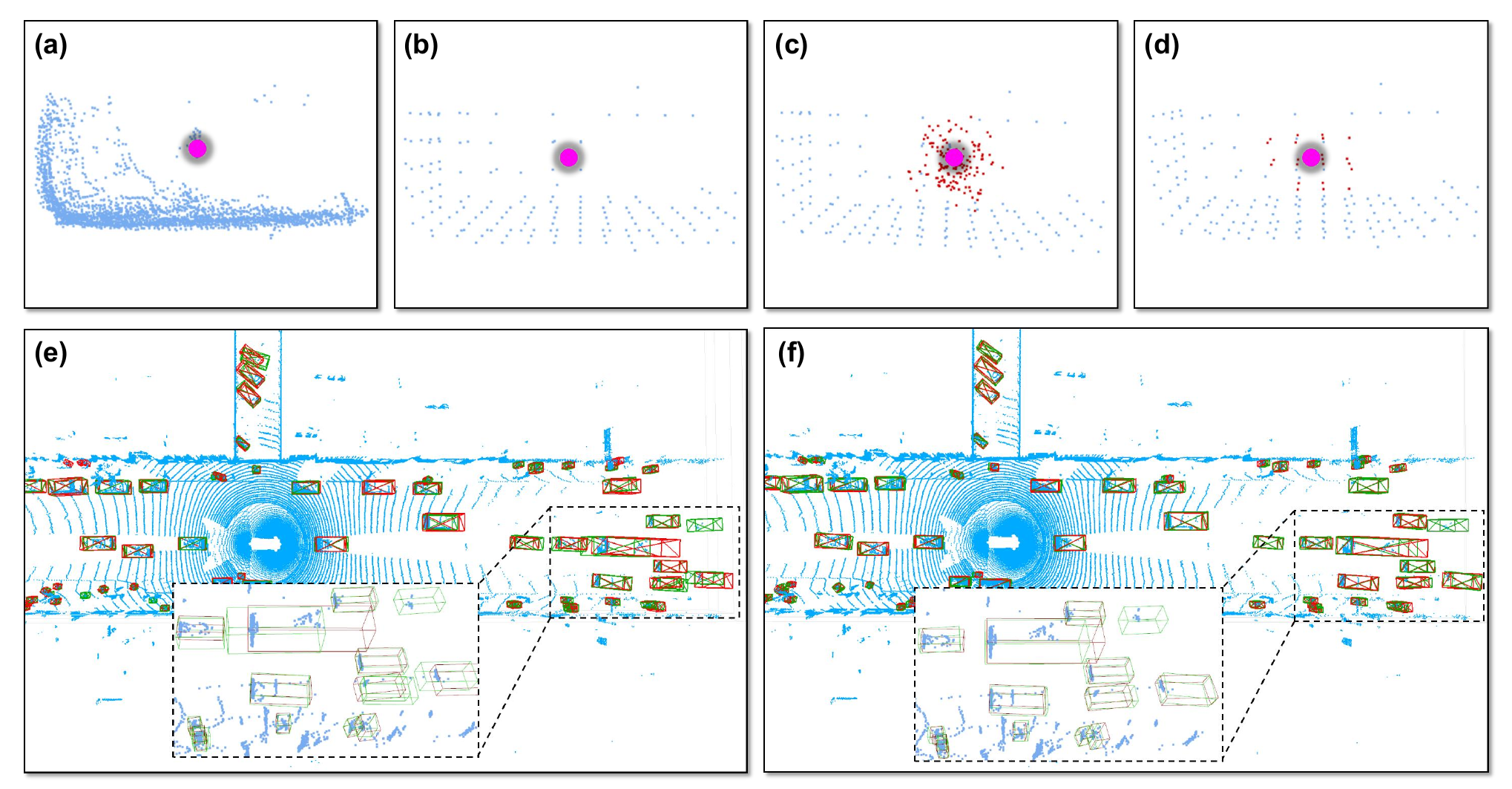}}
  \caption{\textbf{Top:} Intermediate result obtained through the process of filling the center of the object. (a) Raw point cloud for a foreground object, with the object center indicated in \textcolor{cpink}{pink}. (b) Centers of foreground voxels. (c) Generated vote points, highlighted in \textcolor{red}{red}. (d) Centers of newly generated voxels after re-voxelizing the vote points. \textbf{Bottom:} The predicted results (e) with and (f) without the Object Center Voting module. The ground-truths and predictions are represented by \textcolor{red}{red} and \textcolor{green}{green} boxes, respectively. The utilization of the module significantly enhances the accuracy of bounding box localization, particularly for large objects.}
  \label{fig:vis_vote}
\end{figure*}

Moreover, Table~\ref{tab:waymo} exhibits the results of our two-stage detector based on MsSVT++, where a second-stage detection head, namely CT3D~\cite{sheng2021improving}, is incorporated behind our MsSVT++ (SS) model. The resulting two-stage detector, denoted as MsSVT++ (CT3D), achieves superior performance across all categories and significantly surpasses the previous state-of-the-art PV-RCNN++ by $\bm{3.0}$ and $\bm{4.0}$ mAP on the \textit{Pedestrian} and \textit{Cyclist} categories, respectively, using 20\% of the data. Additionally, to evaluate the universality of our method, we also develop a two-stage detector based on MsSVT++ using the LidarRCNN architecture. The resulting two-stage detector, named MsSVT++ (LidarRCNN), exhibits significant improvement over the LidarRCNN baseline, achieving $\bm{5.1}$, $\bm{11.1}$, and $\bm{7.0}$ mAP on the \textit{Vehicle}, \textit{Pedestrian}, and \textit{Cyclist} categories, respectively. These findings highlight the broad applicability of our approach across different architectures. Similarly, Table~\ref{tab:waymo_all} illustrates that our two-stage MsSVT++ (CT3D) model outperforms other models when using 100\% of the available data.

We also provide the runtime comparison between our method and the previous approaches. As depicted in~\cref{table:runtime}, our MsSVT++ (SS) demonstrates superior detection accuracy while preserving a comparable level of inference speed compared to the previous approaches. This validates the efficacy of our method in achieving a commendable accuracy-speed trade-off.

\noindent\textbf{Qualitative Results.}
We present the qualitative results of our approach in~\cref{fig:vis_box}. Notably, our method exhibits precise prediction of bounding boxes even in scenes beyond $50\rm{m}$, characterized by significantly low point density (\cref{fig:vis_box} (a)). This remarkable capability highlights the effectiveness of MsSVT in capturing contextual information, compensating for the absence of fine-grained details in long-range object detection scenarios. Additionally, our model demonstrates impressive performance in complex scenes featuring dense objects with substantial scale variations (\cref{fig:vis_box} (b)), underscoring the flexibility and robustness of MsSVT.

Nevertheless, when dealing with distant, isolated objects (\cref{fig:vis_box} (c)), the points representing the target object exhibit a notably sparse distribution, lacking local geometric information. Furthermore, due to the absence of proximate objects, there is a dearth of longer-range contextual information. In these specific scenarios, our method may encounter occasional difficulties in detecting the target object. This limitation can be primarily attributed to the intrinsic sparsity inherent to point cloud data.

\begin{table*}[t]
\caption{Results on KITTI validation and test set.}
\label{tab:kitti}
\renewcommand{\arraystretch}{1.15}
\setlength{\tabcolsep}{6.0pt}
\centering
\begin{tabular}{l|c|ccc|ccc|ccc|ccc}
    \hline
    \hline
    & & \multicolumn{9}{c|}{\textbf{Validation}}  & \multicolumn{3}{c}{\textbf{Test}} \\ 
    \cline{3-14}  Method & Reference & \multicolumn{3}{c|}{3D Car (IoU=0.7)} &  \multicolumn{3}{c|}{3D Ped. (IoU=0.5)} & \multicolumn{3}{c|}{3D Cyc. (IoU=0.5)} & \multicolumn{3}{c}{3D Car (IoU=0.7)} \\ 
    & & Easy & Mod. & Hard & Easy & Mod. & Hard & Easy & Mod. & \multicolumn{1}{c|}{Hard} & Easy & Mod. & Hard \\

    \hline
    \multicolumn{14}{c}{\bf{Single-Stage Methods}}\\
    \hline
    SECOND~\cite{yan2018second} & Sensors 2018 & 86.46 & 77.28 & 74.65 & 61.63 & 56.27 & 52.60 & 80.10 & 62.69 & 59.71 & 84.65 & 75.96 & 68.71 \\
    PointPillar~\cite{lang2019pointpillars} & CVPR 2019 & 88.61 & 78.62 & 77.22 & 56.55 & 52.98 & 47.73 & 80.59 & 67.16 & 63.11 & 82.58 & 74.31 & 68.99 \\
    3DSSD~\cite{yang20203dssd} & CVPR 2020  & 88.55 & 78.45 & 77.30 & 58.18 & 54.32 & 49.56 & 86.25 & 70.49 & 65.32 & 88.36 & 79.57 & 74.55 \\
    VoTr-SS~\cite{mao2021voxel} & ICCV 2021 & 87.86 & 78.27 & 76.93 & -- & -- & -- & -- & -- & --   & 86.73 & 78.25 & 72.99 \\
    VoxelSet~\cite{he2022voxel} & CVPR 2022 & 88.45 & 78.48 & 77.07 & 60.62 & 54.74 & 50.39 & 84.07 & 68.11 & 65.14   & 88.53 & 82.06 & 77.46 \\
    MsSVT (SS)~\cite{dong2022mssvt} & NeurIPS 2022 & 89.08 & 78.75 & 77.35 & 63.59 & 57.33 & 53.12 & 88.57 & 71.70 & 66.29 & 90.04 & 82.10 & 78.25  \\
    \rowcolor{RowColor} MsSVT++ (SS) & -- & \textbf{89.21} & \textbf{79.24} & \textbf{78.12} & \textbf{64.47} & \textbf{57.62} & \textbf{53.45} & \textbf{89.04} & \textbf{72.45} & \textbf{66.38}  & 90.19 & 82.43 & 78.58 \\
    \hline
    \multicolumn{14}{c}{\bf{Two-Stage Methods}}\\
    \hline
    PointRCNN~\cite{shi2019pointrcnn} & CVPR 2019 & 89.03 & 78.78 & 77.86 & 62.50 & 55.18 & 50.15 & 87.49 & 72.55 & 66.01 & 86.96 & 75.64 & 70.70 \\
    Part-A2~\cite{shi2019part} & TPAMI 2020 & 88.48 & 78.96 & 78.36 & \textbf{70.73} & \textbf{64.13} & \textbf{57.45} & 88.18 & 73.35 & \textbf{70.75}  & 87.81 & 78.49 & 73.51 \\
    PV-RCNN~\cite{shi2020pv} & CVPR 2020 & \textbf{89.35} & 83.69 & 78.70 & 63.12 & 54.84 & 51.78 & 86.06 & 69.48 & 64.50  & \textbf{90.25} & 81.43 & 76.82 \\
    VoTr-TS~\cite{mao2021voxel} & ICCV 2021 & 89.04 & 84.04 & 78.68 & -- & -- & -- & -- & -- & --  & 89.90 & 82.09 & \textbf{79.14} \\
    PV-RCNN++~\cite{shi2023pv} & IJCV 2023 & -- & -- & -- & -- & -- & -- & -- & -- & -- & 90.14 & 81.88 & 77.15 \\
    MsSVT (CT3D)~\cite{dong2022mssvt} & NeurIPS 2022 & 89.32 & 84.66 & \textbf{78.94} & 66.11 & 58.94 & 53.86 & 92.49 & 73.60 & 69.34 & 90.11 & 82.52 & 78.37 \\
    \rowcolor{RowColor} MsSVT++ (CT3D) & -- & 89.24 & \textbf{84.93} & 78.90 & 66.37 & 59.58 & 53.92 & \textbf{92.63} & \textbf{73.98} & 69.31  & 90.22 & \textbf{82.87} & 79.01 \\
    \hline
    \hline
\end{tabular}
\end{table*}

\begin{table*}[t]
\caption{Results on Argoverse 2 validation set. *: implemented by FSD~\cite{fan2023super}.}
\label{tab:argoverse}
\renewcommand{\arraystretch}{1.15}
\setlength{\tabcolsep}{3.5pt}
\centering
\begin{tabular}{l|c|cccccccccccccccccccc}
    \hline
    \hline
    Method & \rotatebox{90}{Average} & \rotatebox{90}{Vehicle} & \rotatebox{90}{Bus} & \rotatebox{90}{Pedestrian} & \rotatebox{90}{Stop Sign} & \rotatebox{90}{Box Truck} & \rotatebox{90}{Bollard} & \rotatebox{90}{C-Barrel} & \rotatebox{90}{Motorcyclist} & \rotatebox{90}{MPC-Sign} & \rotatebox{90}{Motorcycle} & \rotatebox{90}{Bicycle} & \rotatebox{90}{A-Bus} & \rotatebox{90}{School Bus} & \rotatebox{90}{Truck Cab} & \rotatebox{90}{C-Cone} & \rotatebox{90}{V-Trailer} & \rotatebox{90}{Sign} & \rotatebox{90}{Large Vehicle} & \rotatebox{90}{Stroller} & \rotatebox{90}{Bicyclist} \\
    \hline
    \multicolumn{22}{c}{\bf{Precision}}\\
    \hline
    CenterPoint*~\cite{yin2021center} & 22.0 & 67.6 & 38.9 & 46.5 & 16.9 & 37.4 & 40.1 & 32.2 & 28.6 & 27.4 & 33.4 & 24.5 & ~8.7 & 25.8 & 22.6 & 29.5 & 22.4 & ~6.3 & ~3.9 & ~0.5 & 20.1 \\
    FSD~\cite{fan2022fully} & 24.0 & 67.1 & 39.8 & 57.4 & 21.3 & 38.3 & 38.3 & 38.1 & 30.0 & 23.6 & 38.1 & 25.5 & 15.6 & 30.0 & 20.1 & 38.9 & 23.9 & 7.9 & 5.1 & 5.7 & 27.0 \\
    MsSVT (SS)~\cite{dong2022mssvt} & 24.1 & 67.4 & 40.2 & 56.0 & 22.6 & 37.9 & 41.3 & 37.2 & 30.7 & 23.1 & 39.5 & 24.1  & 15.3 & 32.1 & 19.7 & 39.2 & 25.5 & 9.6 & 5.0 & 5.1 & 25.8 \\
    \rowcolor{RowColor} MsSVT++ (SS) & 24.5 & 67.9 & 41.0 & 56.2 & 22.9 & 38.5 & 41.2 & 37.6 & 30.4 & 22.9 & 39.8 & 24.5 & 16.4 & 33.2 & 20.7 & 39.1 & 25.8 & 10.3 & ~5.5 & ~5.0 & 26.1 \\
    \hline
    \multicolumn{22}{c}{\bf{Composite Score}}\\
    \hline
    CenterPoint*~\cite{yin2021center} & 17.6 & 57.2 & 32.0 & 35.7 & 13.2 & 31.0 & 28.9 & 25.6 & 22.2 & 19.1 & 28.2 & 19.6 & ~6.8 & 22.5 & 17.4 & 22.4 & 17.2 & ~4.8 & ~3.0 & ~0.4 & 16.7 \\
    FSD~\cite{fan2022fully} & 19.1 & 56.0 & 33.0 & 45.7 & 16.7 & 31.6 & 27.7 & 30.4 & 23.8 & 16.4 & 31.9 & 20.5 & 12.0 & 25.6 & 15.9 & 29.2 & 18.1 & 6.4 & 3.8 & 4.5 & 22.1 \\
    MsSVT (SS)~\cite{dong2022mssvt} & 18.9 & 56.4 & 33.6 & 44.3 & 16.9 & 31.9 & 25.8 & 27.7 & 24.2 & 16.0 & 32.3 & 17.2 & 11.8 & 26.6 & 14.5 & 28.7 & 19.9 & 7.9 & 3.6 & 4.2 & 20.8 \\
    \rowcolor{RowColor} MsSVT++ (SS) & 19.4 & 57.1 & 34.3 & 44.6 & 18.1 & 36.2 & 24.3 & 29.8 & 24.0 & 15.6 & 33.5 & 17.7 & 12.6 & 28.2 & 15.4 & 28.8 & 20.3 & ~8.4 & ~4.2 & ~4.0 & 21.5 \\
    \hline
    \hline
\end{tabular}
\end{table*}

\noindent\textbf{Visualization of Attention Maps.}
To gain a better understanding of the behavior of our MsSVT, we visualize attention maps that indicate what the model has focused on. As depicted in the upper portion of \cref{fig:vis_weight}, the head group characterized by a smaller key window prioritizes local foreground information, while the head group with a larger key window attends more to longer-range context. Consequently, these two head groups complement each other and effectively capture information at mixed scales, thereby facilitating the detection of objects with diverse scales.

Furthermore, in the lower section of~\cref{fig:vis_weight}, we present additional visualization results of our predicted bounding boxes alongside their corresponding attention weights in challenging cases. As observed, the majority of objects in~\cref{fig:vis_weight} (e) and (f) are detected with high accuracy. Notably, in certain instances indicated by dotted rectangles, the bounding boxes contain only a limited number of points (ranging from 1 to 5). However, despite this sparse point representation, these bounding boxes are still accurately predicted with confidence scores surpassing 0.5.

Upon visualizing the attention weights, we observe that in these challenging cases, the limited number of points within the bounding boxes exhibit heightened attention towards the neighboring objects to gather supplementary contextual information, compensating for the sparse points within the boxes. This observation underscores the ability of MsSVT to capture a wider spectrum of contextual information, enabling accurate inference of the box's location and size even in the absence of intricate details. A representative illustration of this phenomenon can be observed in the lower section of \cref{fig:vis_weight} (e) and (f): within a row of adjacent vehicles, even if one vehicle is scarcely covered by points, its bounding box can still be precisely predicted by leveraging the comprehension of the scene's semantics and analyzing the spatial distribution and dimensions of the surrounding vehicles.

\noindent\textbf{Visualization of Voted Voxels.}
To facilitate a more holistic comprehension of the functionality of our proposed Center Voting module, we present a visualization of the intermediate results obtained during the object center filling process. As depicted in the upper section of \cref{fig:vis_vote}, the foreground voxels of the object effectively generate vote points that exhibit a compact clustering pattern around the object's center. Consequently, this process effectively populates the previously vacant voxel space within the object center with new voxels that have assimilated information from diverse regions of the object.

In the lower portion of \cref{fig:vis_vote}, we present predicted results, specifically (e) with and (f) without the utilization of the Center Voting module. The inclusion of this module significantly enhances the accuracy of bounding box localization, particularly for large objects. For instance, consider the \textit{long bus} located in the top-middle section of the zoomed-in figure. Without utilizing the Center Voting module, the predicted bounding box deviates significantly from the ground-truth. This discrepancy arises due to the fact that the LiDAR points containing crucial structural information are predominantly concentrated at the bus's tail, distant from its center. Consequently, the detection head is compelled to make more aggressive predictions for larger bounding box offset values, which undoubtedly poses a more intricate challenge.

Conversely, the incorporation of the Center Voting module facilitates the aggregation of information originating from diverse regions of the object onto the newly generated voxels that envelop the object center. This consolidation of information simplifies the prediction process based on these augmented voxels, thereby yielding a bounding box that comprehensively encapsulates the entirety of the object and aligns more closely with the ground-truth.

\subsection{Results on KITTI}
\label{sec:kitti}
\noindent\textbf{Setups.} 
Our model is also assessed on the KITTI benchmark~\cite{geiger2013vision}. To prepare the input point cloud, a filtering process is applied to retain points falling within the following ranges: $\left[0\rm{m}, 70.4\rm{m} \right]$ along the $x$-axis, $\left[-40.0\rm{m}, 40.0\rm{m} \right]$ along the $y$-axis, and $\left[-3.0\rm{m}, 1.0\rm{m} \right]$ along the $z$-axis. The voxel size is set to $\left(0.32\rm{m}, 0.32\rm{m}, 0.4\rm{m}\right)$, and all other settings remain constant, as in the experiments conducted on the Waymo dataset.
The model is trained for $100$ epochs using the Adam optimizer with a learning rate of $0.003$, which undergoes cyclic decay~\cite{yan2018second}. The evaluation metric used is the 3D mean average precision (mAP) for three difficulty levels: easy, moderate, and hard.

\noindent\textbf{Main Results.}
\cref{tab:kitti} depicts that our model achieves competitive performance across all three categories on the KITTI validation set. Our single-stage MsSVT++ (SS) outperforms the superior VoTr~\cite{mao2021voxel} by $1.0$ mAP on the \textit{Car} category, and it surpasses some leading two-stage detectors on the \textit{Pedestrian} and \textit{Cyclist} categories. Furthermore, our two-stage MsSVT++ (CT3D) achieves the highest performance on the \textit{Car} and \textit{Cyclist} categories, with mAP values of $84.93$ and $73.98$, respectively, thereby further widening our lead over the competition.
The performance of our method on the KITTI test set shows similar trends. Our single-stage MsSVT++ (SS) outperforms the superior two-stage PV-RCNN++ on the \textit{Car} category, while our two-stage MsSVT++ (CT3D) surpasses the prevailing VoTr-TS by a large margin of $0.8$ mAP. These results convincingly illustrate the generalizability of our detectors based on MsSVT++ across diverse datasets.

\subsection{Results on Argoverse 2}
\noindent\textbf{Setups.} 
Argoverse 2~\cite{wilson2argoverse} is a recently released large-scale dataset consisting of 1000 sensor annotation sequences. This dataset comprises objects from 30 distinct classes and presents the challenge of the long-tail distribution. For our model implementation, we take a single-frame point cloud as input and crop it to a size of 200 m $\times$ 200 m. The model is trained for 30 epochs using the AdamW optimizer with a learning rate of 0.001, following the same configuration as described in~\cite{fan2023super}. In addition to Average Precision (AP), AV2 incorporates Composite Score as an evaluation metric, which accounts for both the average precision (AP) and localization errors.

\noindent\textbf{Main Results.}
In our comparative analysis, we evaluate our model against the methods presented in~\cite{fan2023super}. Follow this work, we exclude the results for the \textit{Dog}, \textit{Wheelchair}, and \textit{Message board trailer} categories due to their limited number of instances. Across all classes, our MsSVT++ (SS) achieves comparable or superior performance compared to other existing detectors. Notably, our MsSVT++ (SS) significantly outperforms the recent state-of-the-art FSD in the detection of small objects, such as the \textit{Sign} category, with improvements of $\bm{2.4}$ and $\bm{2.0}$ in Precision and Composite Score, respectively. Additionally, our MsSVT++ (SS) exhibits better performance than FSD in the detection of extremely large objects, such as the \textit{School Bus} and \textit{V-Trailer}, with improvements of $\bm{3.2}$ and $\bm{2.6}$ in Precision and Composite Score for the \textit{School Bus} category, respectively. These results unequivocally demonstrate the superiority of our method in recognizing and accurately localizing objects across a wide range of scales.

\subsection{Ablation Study}
We conducted thorough ablation studies to validate our design choices and parameter settings. All ablation models were trained for 12 epochs using 20\% of the Waymo dataset. We measure the latency on Tesla-V100 GPU, employing Unbuntu-16.04, Python 3.7, Cuda-10.2, and Pytorch-1.8.

\subsubsection{Effect of Key Components}
We evaluated the efficacy of crucial components, namely the Balanced Multi-window Sampling strategy, scale-aware head attention mechanism, scale-aware relative position encoding, and Center Voting module. We progressively incorporated these components in order of increasing complexity, as depicted in Table~\ref{table:ab_main}.

\noindent\textbf{Balanced Multi-window Sampling.} 
We initially validate the effectiveness of our Balanced Multi-window Sampling strategy. The base model, listed in the first row, employs a 3D version of standard window-based attention~\cite{liu2021swin} with a window size of $(3, 3, 5)$ and without the shift window scheme. It adopts dilated key sampling~\cite{mao2021voxel} and gathers all non-empty voxels within the window as queries. To evaluate the impact of our proposed Balanced Multi-window Sampling, we replace the dilated key sampling with this approach in a model variant. The results in the second row demonstrate that sampling key voxels from multiple windows of different sizes yields noticeable performance gains, increasing from $71.66$ to $73.44$ on \textit{Pedestrian} and from $63.31$ to $66.88$ on \textit{Cyclist}.

\begin{table}[t]
    \caption{Ablations on key components. BMS: Balanced Multi-window Sampling, SHA: Scale-aware Head Attention, SRPE: Relative Position Encoding, CV: Center Voting (Averaged Point Feature), MC: Mixed-scale Contextual Feature.}
    \label{table:ab_main}
    \renewcommand{\arraystretch}{1.2}
    \setlength{\tabcolsep}{7.0pt}
    \centering
    \begin{tabular}{ccccc| c}
        \toprule
        BMS & SHA & SRPE & CV & MC &  Veh / Ped / Cyc \\
        \midrule
        \xmark & \xmark &  \xmark & \xmark & \xmark & $69.51$ / $71.66$ / $63.31$ \\
        \cmark & \xmark &  \xmark & \xmark & \xmark & $71.24$ / $73.44$ / $66.88$ \\
        \cmark & \cmark &  \xmark & \xmark & \xmark & $71.96$ / $75.08$ / $67.16$ \\
        \cmark & \cmark &  \cmark & \xmark & \xmark & $72.37$ / $75.99$ / $67.90$ \\
        \cmark & \cmark &  \cmark & \cmark & \xmark & $73.63$ / $76.53$ / $69.02$ \\
        \cmark & \cmark &  \cmark & \cmark & \cmark & $73.86$ / $76.47$ / $69.18$ \\
        \bottomrule
    \end{tabular}
\end{table}

\begin{figure}[t]
  \centering{\includegraphics[width=0.50\textwidth]{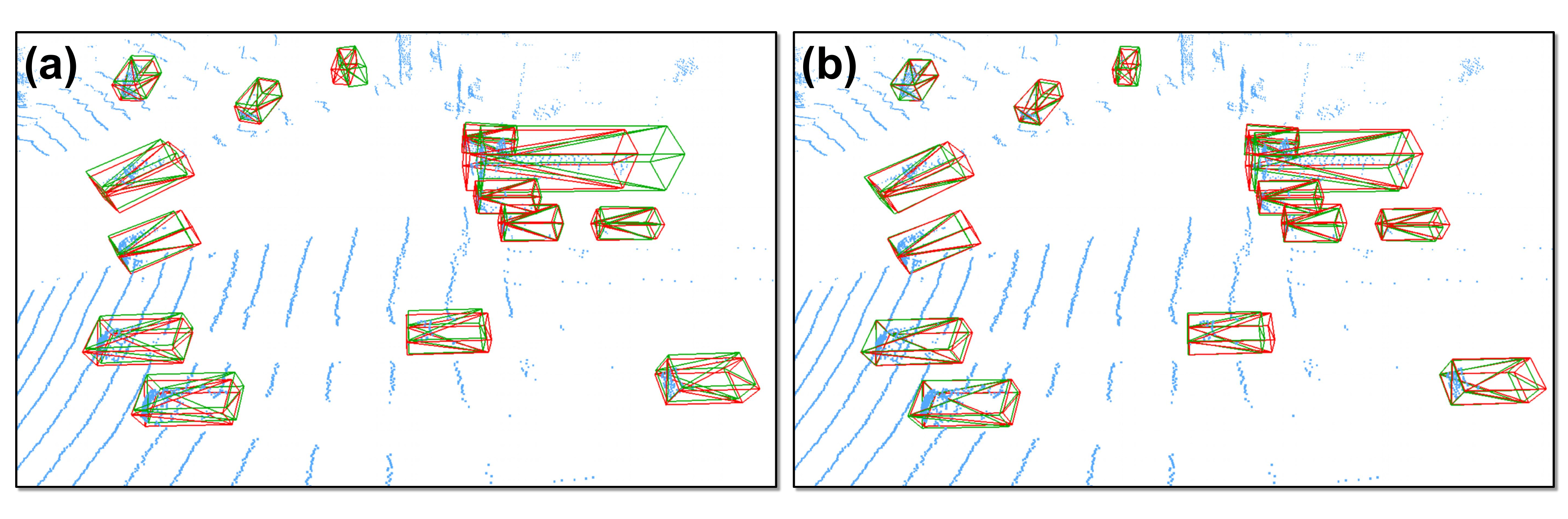}}
  \caption{Comparison of detection outcomes (a) without and (b) with the inclusion of mixed-scale contextual features in the Center Voting module. Ground-truth and predicted bounding boxes are depicted in \textcolor{red}{red} and \textcolor{green}{green}, respectively. The incorporation of mixed-scale contextual features improves object localization accuracy.}
  \label{fig:mixed-scale}
\end{figure}

\noindent\textbf{Scale-aware Head Attention.}
Next, we validate the effectiveness of our proposed scale-aware head attention mechanism. By comparing the results in the second and third rows of~\cref{table:ab_main}, it becomes evident that enabling multiple head groups to capture information at different scales significantly enhances the performance, resulting in an improvement from $73.44$ to $75.08$ on the \textit{Pedestrian} category.

\noindent\textbf{Scale-aware Relative Position Encoding.}
The results in the third and fourth rows of~\cref{table:ab_main} demonstrate that incorporating scale-aware relative position encoding further improves the performance compared to the model variant that utilizes scale-agnostic position encoding. These results strongly support our design motivation, indicating that the position encoding should vary with different scales.

\noindent\textbf{Center Voting Module.}
The results presented in the final three rows of~\cref{table:ab_main} demonstrate that incorporating the voted voxels, which are occupied with averaged point features, to object centers leads to an improvement in AP for all categories. Furthermore, augmenting these voted voxels with mixed-scale contextual features results in a further performance boost. Overall, there is a significant increase in AP of $1.49$ and $1.28$ for the \textit{Vehicle} and \textit{Cyclist} categories, respectively. The improvement for smaller-scale objects, such as \textit{Pedestrian}, is relatively modest at $0.48$, primarily due to the reduced number of empty voxels at object centers. This observation suggests that the Center Voting module enhances the predictive capability of our model, particularly for objects with large sizes.

To enhance the credibility of augmented mixed-scale contextual features, we conducted a comparative analysis of object detection outcomes, assessing their performance with and without the incorporation of these features. As depicted in \cref{fig:mixed-scale}, the utilization of mixed-scale contextual features yields predicted bounding boxes that exhibit a close alignment with the ground-truth bounding boxes. This observation underscores the valuable role played by mixed-scale contextual features in enhancing object localization accuracy. This enhancement results from the concurrent amalgamation of cues stemming from various object parts and the incorporation of longer-range contextual information.

\subsubsection{Impact of Mixed-scale Window Settings}
In this section, we examine the influence of various configurations for key windows. These configurations include the number of scales assigned to key windows, the allocation of head groups for key windows of different scales, and the sizes of the key windows. The size of the query window are consistently set at [3, 3, 5]. To focus solely on evaluating the effectiveness of the Mixed-scale Sparse Voxel Transformer, the ablation studies described below exclude the Center Voting module.

\noindent\textbf{The Number of Scales for Key Windows.}
We have implemented two sets of key windows with distinct scales, specifically [3, 3, 5] and [7, 7, 7]. The primary objective of our investigation is to assess the impact of mixed-scale configurations on the number of key window scales, while maintaining a constant total of 8 attention heads. Initially, we employed key windows that had a single scale, namely [3, 3, 5] or [7, 7, 7]. The outcomes presented in the first two rows of~\cref{table:mix} demonstrate that utilizing a single-scale key window results in lower accuracy due to insufficient modeling of multi-scale information.

In contrast, we introduced additional windows with sizes of [5, 5, 7] and [9, 9, 7] to incorporate more scales. However, the middle three rows demonstrate that the introduction of more scales does not improve accuracy but instead increases runtime latency. This can be explained by our Scale-aware Head Attention mechanism, which assigns attention heads (feature channels) to different scales of windows, allowing each group of heads to handle a specific scale. If the key window scales are too dispersed, it may reduce the number of feature channels available for each head group, potentially reducing the representation capacity of features specific to a particular scale.
Furthermore, the results indicate that introducing and allocating more attention heads to oversized windows, such as [9, 9, 7], has an impact on the accuracy of small objects. The accuracy on the \textit{Cyclist} category decreases from $67.01$ to $66.53$.

\noindent\textbf{The Size of Key Windows.}
By fixing the number of scales for key windows at 2, we conducted further investigations into the impact of window size for each scale, as depicted in~\cref{table:window_size}. Our findings consistently demonstrated a decline in performance with increasing window size. We attribute this decline to the sparsity of key voxels within larger windows, which results in a loss of fine-grained information. One potential solution to address this issue is to increase the number of sampled key voxels. However, this approach comes with a significant computational cost and may not be practical for implementation. Consequently, we decided to set the window size of the two scales to [3, 3, 5] and [7, 7, 7], respectively, in our experiments.

 \begin{table}[t]
\caption{Ablations on mixed-scale window settings.}
\label{table:mix}
\renewcommand{\arraystretch}{1.2}
\setlength{\tabcolsep}{3.0pt}
\centering
\begin{tabular}{ccccc|c|c}
\toprule
\multirow{2}{*}{} & \multicolumn{4}{c|}{Key Window Size} & \multirow{2}{*}{Veh / Ped / Cyc} & \multicolumn{1}{l}{\multirow{2}{*}{Lat (ms)}} \\ 
 & \multicolumn{1}{l}{{[}3,3,5{]}} & \multicolumn{1}{l}{{[}5,5,7{]}} & \multicolumn{1}{l}{{[}7,7,7{]}} & \multicolumn{1}{l|}{{[}9,9,7{]}} &  &  \\
\midrule
\multirow{8}{*}{\rotatebox{90}{Num of Head}} & $\times 8$ & -- & -- & -- & 71.46 / 74.39 / 65.55 & 70 \\
& -- & -- & $\times 8$ & -- & 72.06 / 75.20 / 66.33 & 82 \\
\cline{2-7} & $\times 2$  & $\times 3$  & $\times 3$  & -- & 71.94 / 75.76 / 67.07 & 110 \\
& $\times 2$  & $\times 2$  & $\times 4$  & -- & 72.17 / 75.45 / 67.01 & 111 \\
& $\times 2$  & $\times 2$  & $\times 2$  & $\times 2$  &  72.04 / 75.38 / 66.53 & 128 \\
 \cline{2-7} & $\times 2$  & -- & $\times 6$  & -- & 72.23 / 75.81 / 68.08 & 93 \\
 & $\times 6$  & -- & $\times 2$  & -- & 71.96 / 75.65 / 67.65 & 92 \\
& $\times 4$  & -- & $\times 4$  & -- & 72.37 / 75.99 / 67.90 & 92 \\
\bottomrule                                           
\end{tabular}
\end{table}

\begin{table}[t]
    \caption{Ablations on the size of key windows.}
    \label{table:window_size}
    \renewcommand{\arraystretch}{1.2}
    \setlength{\tabcolsep}{20.0pt}
    \centering
    \begin{tabular}{l|c}
        \toprule
        Window Size & \multirow{2}{*}{Veh / Ped / Cyc} \\
        (Key1 \ + \ Key2) & \\
        \midrule
        {[}5,\ 5,\ 5{]} \ + \ {[}9,\ 9,\ 9{]} & $72.04$ / $75.37$ / $67.75$ \\
        {[}7,\ 7,\ 5{]} \ + \ {[}11,\ 11,\ 10{]} & $70.48$ / $73.84$ / $65.71$ \\
        \rowcolor{RowColor} {[}3,\ 3,\ 5{]} \ + \ {[}7,\ 7,\ 7{]} & $72.37$ / $75.99$ / $67.90$ \\
        \bottomrule
    \end{tabular}
\end{table}

\noindent\textbf{The Allocation of Attention Heads.} 
Following the configuration of key windows with two different scales, we examined the impact of allocating varying numbers of attention heads to each scale. The outcomes, presented in the last three rows of~\cref{table:mix}, suggest that the performance of our model is not significantly influenced by the allocation of different numbers of attention heads to distinct scales.

\begin{table}[t]
    \caption{Ablations on the sampling strategy for acquiring query and keys.}
    \label{table:samp}
    \renewcommand{\arraystretch}{1.2}
    \setlength{\tabcolsep}{11.0pt}
    \centering
    \begin{tabular}{c|c|c}
        \toprule
        Strategy & \multirow{2}{*}{Veh / Ped / Cyc} & \multirow{2}{*}{Lat (ms)} \\
        (Query \ + \ Key) & \\
        \midrule
        FPS \ + \ FPS & $72.02$ / $75.57$ / $67.91$ & 115 \\
        CBS \ + \ CBS & $71.83$ / $75.28$ / $67.25$ & 78 \\
         FPS \ + \ CBS & $71.88$ / $75.04$ / $67.33$ & 89 \\
         \rowcolor{RowColor} CBS \ + \ FPS & $72.37$ / $75.99$ / $67.90$ & 92 \\
        \bottomrule
    \end{tabular}
\end{table}

\begin{table}[t]
    \caption{Ablations on the sampling rate of Chessboard Sampling.}
    \label{table:sampling}
    \renewcommand{\arraystretch}{1.2}
    \setlength{\tabcolsep}{5.0pt}
    \centering
    \begin{tabular}{c|c|cc}
        \toprule
        Sampling Rate & Veh / Ped / Cyc & Mem (G) & Lat (ms) \\
        \midrule
        w/o & $72.58$ / $75.74$ / $68.24$ & 18.0 & 142 \\
        1/2 & $72.44$ / $76.03$ / $67.81$ & 14.2 & 110 \\
        1/8 & $72.01$ / $75.54$ / $67.43$ & 11.1 & 88 \\
        \rowcolor{RowColor}  1/4 & $72.37$ / $75.99$ / $67.90$ & 11.8 & 92 \\
        \bottomrule
    \end{tabular}
\end{table}

\subsubsection{Impact of Voxel Sampling Settings}
\noindent\textbf{Sampling Strategy for Acquiring Query and Keys.}
We utilize Chessboard Sampling (CBS) and FPS to sample voxels in the query and key window, respectively. To investigate alternative sampling strategies, additional experiments are conducted, and the outcomes are presented in~\cref{table:samp}. In comparison to the CBS (query) + FPS (key) approach, using FPS for both queries and keys leads to increased latency and a slight performance penalty on the \textit{Vehicle} and \textit{Pedestrian} categories. Conversely, employing CBS exclusively results in faster processing but diminished accuracy. The CBS (query) + FPS (key) combination strikes a balance between latency and accuracy by leveraging the strengths of each sampling method. CBS substantially reduces the computational cost associated with query sampling while mitigating information loss arising from uncertain voxel updating. Meanwhile, FPS better preserves the geometric structure. Consequently, this combined approach achieves a favorable trade-off between accuracy and speed.

\noindent\textbf{Sampling Rate of Chessboard Sampling.}
We aim to investigate the impact of utilizing different sampling rates for Chessboard Sampling. As depicted in~\cref{table:sampling}, our model demonstrates robustness across various sampling rates. Notably, when no sampling is applied, the highest accuracy is achieved. However, this comes at the expense of a substantial memory footprint and slower speed. In contrast, incorporating Chessboard Sampling with a sampling rate of 1/4 results in only minor performance degradation compared to the unsampled model variant. Furthermore, this approach significantly enhances computational efficiency by reducing both memory footprint (up to $\bm{33\%}$) and latency (up to $\bm{28\%}$).

\subsubsection{Impact of the Number of MsSVT Blocks}
The proposed backbone network, MsSVT, consists of multiple MsSVT blocks. The influence of varying the number of blocks on the model's performance is presented in \cref{table:blocks}. Overall, our model exhibits robustness when the number of blocks is altered. Increasing the number of blocks leads to improved accuracy for \textit{Vehicle}, while the accuracy for \textit{Pedestrian} and \textit{Cyclist} initially rises but subsequently declines. Notably, augmenting the number of blocks results in larger receptive fields, which facilitates the extraction of higher-level semantic information. This proves particularly advantageous for larger objects such as vehicles. To strike a balance between performance and efficiency considerations, we have chosen 4 blocks as the default configuration for our MsSVT network.

\begin{table}[t]
    \caption{Ablations on the number of MsSVT blocks.}
    \label{table:blocks}
    \renewcommand{\arraystretch}{1.2}
    \setlength{\tabcolsep}{7.0pt}
    \centering
    \begin{tabular}{c|c|cc}
        \toprule
        Blocks & Veh / Ped / Cyc & Mem (G) & Lat (ms) \\
        \midrule
        2 & $71.79$ / $74.66$ / $66.36$ & 8.1 & 71\\
        3 & $72.06$ / $75.24$ / $66.41$ & 10.0 & 80 \\
        4 & $72.37$ / $75.99$ / $67.90$ & 11.8 & 92 \\
        5 & $72.69$ / $75.74$ / $66.88$ & 13.7 & 109 \\
        \bottomrule
    \end{tabular}
\end{table}

\begin{table}[t]
    \caption{Extension to multi-frame point cloud input. 1F: Single-frame, 3F: Multi-frame (3 frames).}
    \label{table:multi_frame}
    \renewcommand{\arraystretch}{1.2}
    \setlength{\tabcolsep}{18.0pt}
    \centering
    \begin{tabular}{l|c}
        \toprule
        Method & Veh / Ped / Cyc \\
        \midrule
        PointPillars-3F~\cite{lang2019pointpillars} & $74.79$ / $73.82$ / $66.95$ \\
        RSN-3F~\cite{sun2021rsn} & $78.40$ / $79.00$ / \ $--$ \ \ \\
        SST-3F~\cite{fan2021embracing} & $77.04$ / $82.42$ / \ $--$ \ \ \\
        \hline
        MsSVT++ (SS-1F) & $78.96$ / $80.64$ / $75.98$ \\
        \rowcolor{RowColor} MsSVT++ (SS-3F) & $79.83$ / $82.26$ / $77.28$ \\
        \bottomrule
    \end{tabular}
\end{table}

\subsubsection{Extension to Multi-frame Input}
The MsSVT++, which employs normal voxels for both input and output, seamlessly supports multi-frame point cloud input. We extend our MsSVT++ (SS) to accommodate multiple frames (3 frames) using the SST approach~\cite{fan2021embracing}. To assess the effectiveness of our extended model, we compare it with several baseline methods that utilize multi-frame point clouds as input, including PointPillar-3F~\cite{lang2019pointpillars}, RSN-3F~\cite{sun2021rsn}, and SST-3F~\cite{fan2021embracing}. Consistent with comparative methods, we underwent training on 100\% of the Waymo dataset and present the LEVEL\_1 AP on the validation split.

As depicted in \cref{table:multi_frame}, the expansion of our model from single-frame input to multi-frame input has led to a substantial enhancement in performance across all categories. Specifically, our results in the Vehicle, Pedestrian, and Cyclist categories have achieved respective accuracies of $79.83\%$, $82.26\%$, and $77.28\%$. This notable performance surpasses PointPillars-3F and RSN-3F, and is on par with SST-3F. These results underscore the scalability of our proposed approach.

\section{Conclusion}
\label{sec:conclusion}
This paper addresses the critical task of 3D object detection from point clouds in large-scale outdoor scenes. We propose a novel approach, namely MsSVT++, which harnesses the capabilities of Mixed-scale Sparse Voxel Transformers to effectively capture both long-range context and fine-grained details. By explicitly partitioning the transformer heads into distinct groups responsible for processing voxels sampled from windows of specific sizes, our divide-and-conquer methodology successfully addresses the challenge of integrating diverse scales of information. The incorporation of the Chessboard Sampling strategy, coupled with the utilization of sparsity and hash mapping techniques, significantly mitigates computational complexities of applying transformers in 3D voxel space. Furthermore, the introduction of the Center Voting module exploits mixed-scale contextual information to populate empty voxels located at object centers, thereby greatly improving object localization, particularly for larger objects. Extensive experimental evaluations validate the superior performance of MsSVT++ in detecting objects across various scales and granularities.

\noindent{\textbf{Limitations.}} 
MsSVT++ has exhibited a commendable ability to effectively capture mixed-scale information by utilizing multiple local windows, demonstrating promising performance. However, a notable limitation is the manual pre-setting of window sizes. In our future endeavors, we intend to explore an adaptive-window variant of MsSVT++ to address this issue.

Furthermore, our MsSVT++ occasionally encounters challenges when addressing remote, isolated objects in extensive point cloud scenes. This limitation can be predominantly attributed to the inherent sparsity of point cloud data. A feasible solution entails integrating supplementary data modalities, such as images, into MsSVT++, thereby providing additional cues to enhance object detection. We also regard the aforementioned proposal as a promising avenue for future research.

\bibliographystyle{IEEEtran}
\bibliography{MsSVT++}

\vfill

\end{document}